\documentclass[11pt]{article}

\usepackage{tabularx}
\usepackage{diagbox}

\usepackage{amsmath,amssymb,graphicx,color,amsthm}
\usepackage[linesnumbered,ruled]{algorithm2e}
\usepackage{tikz}
\usepackage{dsfont}
\usepackage{mathtools}
\usetikzlibrary{shapes.geometric, arrows}
\usepackage{algorithmic}
\usepackage[english]{babel}
\usepackage[utf8]{inputenc}
\usepackage{mathrsfs}  
\usepackage{afterpage}
\usepackage{mathtools}
\usepackage{amsfonts}
\usepackage[colorinlistoftodos]{todonotes}
\usepackage{enumerate}
\usepackage[round]{natbib}
\usepackage{hyperref}
\newcommand{\excise}[1]{}

\newcommand\<{\langle}

\newcommand\RR{\mathbb{R}}

\newcommand\ZZ{\mathbb{Z}}
\newcommand\EE{\mathbb{E}}

\newcommand\tdA{\widetilde{A}}
\newcommand\tdB{\widetilde{B}}
\newcommand\tdE{\widetilde{E}}
\newcommand\tdF{\widetilde{F}}

\newcommand{\tr}{\operatorname{tr}}

\renewcommand\>{\rangle}

\newtheorem{theorem}{Theorem}
\newtheorem{definition}{Definition}
\newtheorem{lemma}{Lemma}

\newtheorem*{example*}{Example}

\DeclareMathOperator\eig{eig}

\DeclarePairedDelimiterX{\infdivx}[2]{(}{)}{%
	#1\;\delimsize\|\;#2%
}

\DeclareMathOperator\stf{St}
\DeclareMathOperator\proj{Proj}
\DeclareMathOperator\sym{Sym}
\DeclareMathOperator{\Cov}{Cov}
\DeclareMathOperator\grad{grad}
\DeclareMathOperator{\Hess}{Hess}
\DeclareMathOperator\id{I}

\usepackage{tikz}
\newcommand*\circled[1]{\tikz[baseline=(char.base)]{
		\node[shape=circle,draw,inner sep=0.5pt] (char) {#1};}}

\newcommand{\RNum}[1]{\uppercase\expandafter{\romannumeral #1\relax}}

\xdefinecolor{dukeblue}{rgb}{0.004,0.129,0.412}

\begin{document}

\title{\mbox{}\\[-11ex]Contrastive inverse regression for dimension reduction}
\author{Sam Hawke$^1$, Hengrui Luo$^2$ and Didong Li$^1$\\ 
	{\em Department of Biostatistics, University of North Carolina at Chapel Hill$^1$}\\
	{\em Computational Research Division, Lawrence Berkeley Laboratory$^2$}}
\date{\vspace{-5ex}}	
\maketitle

\begin{abstract}
Supervised dimension reduction (SDR) has been a topic of growing interest in data science, as it enables the reduction of high-dimensional covariates while preserving the functional relation with certain response variables of interest. However, existing SDR methods are not suitable for analyzing datasets collected from case-control studies. In this setting, the goal is to learn and exploit the low-dimensional structure unique to or enriched by the case group, also known as the foreground group. While some unsupervised techniques such as the contrastive latent variable model and its variants have been developed for this purpose, they fail to preserve the functional relationship between the dimension-reduced covariates and the response variable. In this paper, we propose a supervised dimension reduction method called contrastive inverse regression (CIR) specifically designed for the contrastive setting. CIR introduces an optimization problem defined on the Stiefel manifold with a non-standard loss function. We prove the convergence of CIR to a local optimum using a gradient descent-based algorithm, and our numerical study empirically demonstrates the improved performance over competing methods for high-dimensional data. 
\end{abstract}

\section{Introduction}
\label{sec:intro}
\citet{vogelstein2021supervised} have highlighted the increasing importance of supervised dimension reduction (SDR) methodologies compared to unsupervised counterparts (e.g., principal component analysis (PCA), spherelets \citep{li2022efficient}, spherical rotation component analysis~\citep{luo_li_SPCA_2021}, etc.) in data science. With the advent of large biological datasets, such as single-cell RNA sequencing data, dimension reduction (DR) has become a crucial step in the preprocessing of data for visualization, structural discovery, and downstream biological analyses.

Given paired observations $(x,y)\in \RR^p\times\RR$, where $x$ consists of $p$ covariates, and $y$ is the corresponding response or output variable, the common assumption in SDR is that 
\begin{equation}\label{eqn:assumption}
	y=\varphi(V^\top x,\epsilon),\text{ for some function }\varphi,
\end{equation}
where $V\in\RR^{p\times d}$ with $d\ll p$ is the projection matrix from a high-dimensional to a low-dimensional space, $\epsilon$ is the measurement error independent of $x$, and $\varphi$ is an arbitrary unknown function. For example, in a single-cell RNA sequencing dataset, $x$ could be the expression of genes for a cell and $y$ could be the cell type. 

Under assumption \eqref{eqn:assumption}, although the low-dimensional representation $V^\top x$ is determined by a linear transformation, the function $\varphi$ is an arbitrary unknown function. In this paper, we stick to the assumption in \eqref{eqn:assumption} to focus on linear DR methods for two reasons. First, linear methods are computationally more efficient, particularly for large $p$ and large $n$. Secondly, linear methods are more interpretable, which is an essential characteristic in scientific applications. For instance, in the example above, each column of $V^\top x$ is often considered as a genetic pathway~\cite{bader2006pathguide}. Although our proposed method can be extended to nonlinear cases by kernel method, we will leave this as future work.

Slice Inverse Regression (SIR, \citet{li1991sliced}) is a well-established technique for 
supervised dimension reduction that is widely applicable in multiple scenarios due to its roots in regression analysis. It has been shown to have strong consistency results in both fixed dimensional \citep{hsing1992asymptotic} and high-dimensional  \citep{lin2018consistency} settings. The goal of SIR is to capture the most relevant low-dimensional linear subspace without any parametric or nonparametric model-fitting process for $\varphi$. 

Moreover, SIR offers a geometric interpretation by conditioning on the sufficient statistics of the input distribution~\citep{li1991sliced, cook1991sliced}. SIR incorporates the idea of linear dimension reduction with statistical sufficiency. In SIR, given a pair of features $x\in\mathbb{R}^p$ and univariate response $y\in\mathbb{R}$, the goal is to find a matrix $V\in \mathbb{R}^{p\times d},d<p$ such that $y$ is conditionally independent of $x$ given $V^\top x$. Although the matrix $V$ is not identifiable, the column space of $V$, denoted $\mathcal{C}(V)$, is identifiable. 

Motivated by emerging modern high-dimensional \citep{girard2022advanced, liaoluoma2023_tune, wilkinson_distance-preserving_2022} and biological datasets \citep{hilafu2022sparse,li2008sliced}, SIR evolved and admitted several generalizations, including localized SIR~\citep{wu2008localized}, kernel SIR~\citep{wu2013kernel}, SIR with regularization~\citep{li2008sliced}, SIR for longitudinal data~\citep{jiang2014inverse,li2007tobit}, metric response values~\citep{virta2022sliced}, and online SIR~\citep{cai2020online}. 

In this article, we focus on a specific type of high-dimensional biological data, where the dataset consists of two groups — a foreground group, also known as treatment group or case group, and a background group, also known as control group. The goal is to identify low-dimensional structure, variation, and information unique to the foreground group for downstream analysis. This situation arises naturally in many scientific experiments with two subpopulations. For example, in Electronic Health Record (EHR) data, the foreground data could be health-related covariates from patients who received certain medical treatment, while the background data could be measurements from healthy patients who did not receive any treatment. In this case, the goal is to identify a unique structure in patients who received the treatment that can predict future outcomes. In a genomics context, the foreground data could be gene expression measurements from patients with a disease, and the background data could be measurements from healthy people. In this case, the goal is to predict a certain phenotype for the diseased patient for disease analysis and future therapy.


Previous contrastive models, such as the contrastive latent variable model (CLVM, \citet{zou2013contrastive}),~contrastive principal component analysis (CPCA,~\citet{abid2018exploring}), probabilistic contrastive principal component analysis (PCPCA,~\citet{
	li2020probabilistic}), and the contrastive Poisson latent variable model (CPLVM,~\citet{jones2022contrastive}), have shown that using the case-control structure between foreground and background groups can greatly improve the effectiveness of dimension reduction over standard DR methods such as PCA and its variants. However, to the best of our knowledge, none of these unsupervised contrastive dimension reduction methods is directly applicable to SDR setting.

In this work, we move from these unsupervised contrastive dimension reduction methods to a supervised contrastive dimension reduction setting. By combining the idea of contrastive loss function and the sufficient dimension reduction considered in the SIR model, we propose the Contrastive Inverse Regression (CIR) model, which exactly recovers SIR when a certain parameter is zero. The CIR model sheds light on how to explore and exploit the contrastive structures in supervised dimension reduction. 

\begin{table}[h!]
	
	\centering
	\caption{\label{tab:DRs}Categorization of DR methods by whether they are supervised or contrastive.}
	\begin{tabular}{c|c|c} 
		\hline
		\diagbox[width=12em]{\small Contrastive}{\small Supervised}& No & Yes \\
		\hline
		No & PCA, CCA & SIR, LDA, LASSO  \\
		\hline
		Yes & CPCA, PCPCA & \bf{CIR}\\
		\hline
		
	\end{tabular}
	
\end{table}

Table \ref{tab:DRs} lists several popular DR methods and their properties. The table categorizes these methods as ``contrastive" and ``supervised", based on whether they are designed for case-control data and able to identify low-dimensional structure unique to the case group, and if they take the response variable $y$ into consideration and use $V^\top x$ to predict $y$. 
For example, PCA, the most well known DR method, is neither contrastive nor supervised. Similarly, canonical correlation analysis (CCA,~\citet{hotelling1992relations}) does not utilize $y$ or the unique information of one group, which makes it neither contrastive nor supervised. Methods such as CLVM, CPCA, PCPCA, and CPLVM are contrastive but not supervised. On the other hand, classical supervised DR methods including SIR \citep{li1991sliced}, linear discriminant analysis (LDA,~\citet{hastie1994flexible}), and the least absolute shrinkage and selection operator (LASSO,~\citet{tibshirani1996regression}) are supervised but not contrastive. Our proposed method, CIR, combines both contrastive and supervised features by utilizing both response $y$ and the case-control structure. 

It is important to note that the assumption \eqref{eqn:assumption} does not limit the response variable $y$ to be continuous or categorical, and thus we do not distinguish between regression and classification. However, some methods listed in Table \ref{tab:DRs} are specifically designed for either continuous $y$ (regression, LASSO) or categorical $y$ (classification, LDA). CIR handles both scenarios with the only difference being in the choice of slices, as explained in Section \ref{sec:method}. Furthermore, not all existing DR methods are included in this table. For example, the recently proposed linear optimal low-rank projection (LOL,~\citet{vogelstein2021supervised}) is designed for the classification setting and requires the number of classes to be smaller than the reduced dimension $d$. This can be restrictive, for example, when applied to a single-cell RNA sequencing dataset, where $d$ is required to be greater than number of cell types. In contrast, CIR does not require such data-dependent constraints on the reduced dimension $d$. Similarly, data visualization algorithms that require $d=2$ such as the t-distributed stochastic neighbor embedding (tSNE,~\citet{van2008visualizing}) and uniform manifold approximation and projection (UMAP,~\citet{becht2019dimensionality}) are not listed in the table.

We now present our proposed methodology, including an algorithm for solving the associated nonconvex optimization problem on the Stiefel manifold. We also provide analysis of the convergence of the algorithm, and conduct extensive experiments to demonstrate its superior performance on high-dimensional biomedical datasets when compared to existing DR methods. All proofs and additional experimental details are provided in the appendix.

\section{Method}
\label{sec:method}
To maintain consistency, we will use the terms ``foreground  group " and ``background group" instead of ``case-control" or ``treatment-control" in the remaining sections. We first briefly review SIR as our motivation.

\begin{definition}\label{def:Stiefel} (Stiefel manifold)
	$\stf(p,d)\coloneqq \{V\in \RR^{p\times d}: V^\top V = \id_d\}$ admits a smooth manifold structure equipped with a Riemannian metric, called the Stiefel manifold.  
\end{definition}

Recall that under the assumption in Equation \eqref{eqn:assumption}, the centered inverse regression curve, $\EE[x|y]-\EE[x]$, lies exactly in the linear space spanned by columns of $\Sigma_{xx} V$, denoted by $\mathcal{C}(\Sigma_{xx} V)$, where $\Sigma_{xx}$ is the covariance matrix of $x$. This linear subspace is called the \emph{effective dimension reduced} (e.d.r.) space~\citep{li1991sliced}. As a result, the objective of SIR is to minimize the expected squared distance between $\EE[x|y]$ and $\mathcal{C}(\Sigma_{xx} V)$: 
\begin{equation}\label{eqn:loss_SIR}
	\min_{V\in\stf(p,d)} \EE_y\left[d^2(\EE[x|y]-\EE[x],\mathcal{C}(\Sigma_{xx} V)\right]
\end{equation} where $d$ is the Euclidean distance.

In the contrastive setting, denote foreground data by $(x,y)\in\RR^{p} \times \RR$ and background data by $(\widetilde{x},\widetilde{y})\in\RR^{p}\times\RR$. For convenience, we assume that $x$ and $\widetilde{x}$ are centered at the origin so that $\mathbb{E}[x] = \mathbb{E}[\widetilde{x}] = 0$. Our goal is to find a low-dimensional representation of $x$, denoted by $V^\top x$, such that $y$ is determined by $V^\top x$ while $\widetilde{y}$ is not determined by $V^\top\widetilde x$. The goal of CIR is to find a low-dimensional subspace represented by $V$ such that
\begin{align}\label{eqn:assumption_contrastive}
	\begin{cases}
		y = \varphi(V^\top x,\epsilon)\\
		\widetilde{y} \neq \varphi(V^\top x,\widetilde{\epsilon})
	\end{cases}
\end{align}
for some unknown function $\varphi$.
That is, the column space of $V$ captures the low-dimensional information unique to the foreground group so that we can use $V^\top x$ to predict $y$. 
Instead of optimizing a single loss similar to SIR, CIR aims at optimizing the subspace $\mathcal{C}(\Sigma_{xx}V)$ to minimize the ``constrastive loss function":
\begin{align}
	f(V) &\coloneqq \mathbb{E}_{y}\left[d^2(\mathbb{E}[x~\vert~y],\mathcal{C}\left(\Sigma_{xx}V\right)\right]-\alpha \mathbb{E}_{\widetilde{y}}\left[d^2(\mathbb{E}[\widetilde{x}~\vert~\widetilde{y}] , \mathcal{C}\left(\Sigma_{\widetilde{x}\widetilde{x}}V\right)\right] \label{eqn:cir_loss},
\end{align}
where $\alpha\geq 0$, $\Sigma_{xx} = \Cov (X) \text{ and } \Sigma_{\widetilde{x}\widetilde{x}} = \Cov (\widetilde{X}) \in \mathbb{R}^{p \times p}$, and $d$ is the Euclidean distance. Define the following notation: $$v_y = \mathbb{E}[x\ \vert\ y],~ v_{\widetilde{y}} = \mathbb{E}[\widetilde{x}\ \vert\ \widetilde{y}] \in \mathbb{R}^p$$ $$\Sigma_x = \Cov (v_y),~\Sigma_{\widetilde{x}} = \Cov (v_{\widetilde{y}}) \in \mathbb{R}^{p \times p}$$

$v_y$ (and $v_{\widetilde{y}}$) are called the centered inverse regression curve
\citep{li1991sliced,virta2022sliced}. 
The resulting loss function $f$ balances the effectiveness of dimension reduction between the foreground and background groups. We can adjust the hyperparameter $\alpha$ to express our belief in the importance of the background group. Note that the parameter $\alpha$ appears naturally in other contrastive DR methods, including CPCA and PCPCA.

We need to simplify the loss function $f(V)$ for subsequent analyses. Recall that the projection to the subspace $\mathcal{C}(\Sigma_{xx}V)$ and $\mathcal{C}(\Sigma_{\widetilde{x}\widetilde{x}}V)$ is given by the following projection matrices:
\begin{align*}
	P_{\Sigma_{xx}V} &\coloneqq \Sigma_{xx}V \left[V^\top \Sigma_{xx}^2 V\right]^{-1} V^\top \Sigma_{xx}\\
	P_{\Sigma_{\widetilde{x}\widetilde{x}}V} &\coloneqq\Sigma_{\widetilde{x}\widetilde{x}}V \left[V^\top \Sigma_{\widetilde{x}\widetilde{x}}^2 V\right]^{-1} V^\top \Sigma_{\widetilde{x}\widetilde{x}}.
\end{align*}
Because projection matrices are idempotent, that is, $P_{\Sigma_{xx}V}^2 = P_{\Sigma_{xx}V}$ and  $P_{\Sigma_{\widetilde{x}\widetilde{x}}V}^2 = P_{\Sigma_{\widetilde{x}\widetilde{x}}V}$, 
we can rewrite the loss function as follows: 
\begin{align*}
	f(V) &= \mathbb{E}_{y}\left[d^2(\mathbb{E}[x\ \vert\ y],\mathcal{C}\left(\Sigma_{xx}V\right))\right]- \alpha \mathbb{E}_{\widetilde{y}}\left[d^2(\mathbb{E}[\widetilde{x}\ \vert\ \widetilde{y}] ,\mathcal{C}\left(\Sigma_{\widetilde{x}\widetilde{x}}V\right))\right]\\
	&= \mathbb{E}_y\left[\left\|v_y - P_{\Sigma_{xx}V}v_y\right\|^2\right] - \alpha \mathbb{E}_{\widetilde{y}}\left[\left\|v_{\widetilde{y}} - P_{\Sigma_{\widetilde{x}\widetilde{x}}V} v_{\widetilde{y}}\right\|^2\right]\\
	&= \mathbb{E}_y\left[v_y^\top v_y - v_y^\top P_{\Sigma_{xx}V}^2 v_y\right] - \alpha \mathbb{E}_{\widetilde{y}}\left[v_{\widetilde{y}}^\top v_{\widetilde{y}} - v_{\widetilde{y}}^\top P_{\Sigma_{\widetilde{x}\widetilde{x}}V} v_{\widetilde{y}}\right]\\
	& = \mathbb{E}_y\left[v_y^\top v_y - v_y^\top P_{\Sigma_{xx}V} v_y\right] - \alpha \mathbb{E}_{\widetilde{y}}\left[v_{\widetilde{y}}^\top v_{\widetilde{y}} - v_{\widetilde{y}}^\top P_{\Sigma_{\widetilde{x}\widetilde{x}}V} v_{\widetilde{y}}\right]\\
\end{align*}
The solution to the optimization problem defined by this loss function, if it exists, leads to our CIR model. 

We remove the constant terms $\EE_y[v_y^\top v_y]$ and $\EE_{\widetilde{y}}[v_{\widetilde{y}}^\top v_{\widetilde{y}}]$ that are independent of $V$ and continue to simplify $f(V)$:
\begin{align*}
	f(V) &= -\mathbb{E}_y[v_y^\top P_{\Sigma_{xx}V} v_y] + \alpha \mathbb{E}_{\widetilde{y}}[v_{\widetilde{y}}^\top P_{\Sigma_{\widetilde{x}\widetilde{x}}V} v_{\widetilde{y}}]\\
	&=  -\mathbb{E}_y[\tr (v_y^\top P_{\Sigma_{xx}V} v_y)] + \alpha \mathbb{E}_{\widetilde{y}}[\tr (v_{\widetilde{y}}^\top P_{\Sigma_{\widetilde{x}\widetilde{x}}V} v_{\widetilde{y}})]\\
	&= -\tr(\Sigma_x P_{\Sigma_{xx}V}) + \alpha \tr (\Sigma_{\widetilde{x}} P_{\Sigma_{\widetilde{x}\widetilde{x}}V})\\
	&=  -\tr \left(V^\top \Sigma_{xx}\Sigma_x \Sigma_{xx} V\left[V^\top \Sigma_{xx}^2 V\right]^{-1} \right) +\alpha \tr\left(V^\top \Sigma_{\widetilde{x}\widetilde{x}}\Sigma_{\widetilde{x}}\Sigma_{\widetilde{x}\widetilde{x}} V\left[V^\top \Sigma_{\widetilde{x}\widetilde{x}}^2 V\right]^{-1}  \right)\\
	&=-\tr(V^\top AV(V^\top BV)^{-1})+\alpha\tr(V^\top \widetilde{A}V(V^\top \widetilde{B}V)^{-1}),
\end{align*}
where $A = \Sigma_{xx}\Sigma_x\Sigma_{xx}$, $B=\Sigma_{xx}^2$, $\widetilde{A} = \Sigma_{\widetilde{x}\widetilde{x}}\Sigma_{\widetilde{x}}\Sigma_{\widetilde{x}\widetilde{x}}$, and $\widetilde{B}=\Sigma_{\widetilde{x}\widetilde{x}}^2$.

Thus, in the case where $\alpha = 0$, CIR reduces to SIR. In this case, the problem can be reparameterized by $V^* = B^{1/2} V$ so that the columns are orthogonal, which reduces the loss function to a quadratic form, yielding a closed-form solution (a generalized eigenproblem). In the case where $\alpha > 0$, however, we cannot perform the same trick for both $B$ and $\widetilde{B}$, so we must resort to numerical approximations. We adopt gradient-based optimization algorithms on $\stf(p,d)$, which are based on the gradient of $f$ given by the following lemma.

\begin{lemma}\label{lem:grad}
	The gradient of $f$ is given by 
	\begin{align*}
		-\frac{1}{2}\grad f(V) &= AV(V^\top BV)^{-1}-BV (V^\top BV)^{-1} V^\top AV (V^\top BV)^{-1}\\
		&~~~~ -\alpha\left(\widetilde{A}V(V^\top \widetilde{B}V)^{-1}-\widetilde{B}V (V^\top \widetilde{B}V)^{-1} V^\top \widetilde{A}V (V^\top \widetilde{B}V)^{-1}\right).
	\end{align*}
\end{lemma}

Note that the gradient $\grad f$ is different from the standard gradient in Euclidean space, denoted by $Df = \frac{\partial f}{\partial V}$. The difference is that $\grad f$ lies in the tangent space of $\stf(p,d)$ at $V$, while the Euclidean version may escape from the tangent space.

\begin{theorem}\label{thm:critical}
	If $V$ is a local minimizer of the optimization problem \eqref{eqn:cir_loss}, then
	\begin{align*}
		AVE(V)-\alpha \widetilde{A}V\widetilde{E}(V) = BVF(V)-\alpha \widetilde{B}V\widetilde{F}(V),
	\end{align*}
	where $E(V) = (V^\top BV)^{-1}$, $\tdE(V) = (V^\top \tdB V)^{-1}$, $F(V) = (V^\top BV)^{-1} V^\top AV (V^\top BV)^{-1}$, and $\tdF(V) = (V^\top \tdB V)^{-1} V^\top \tdA V (V^\top \tdB V)^{-1}$.
\end{theorem}

Let $G(V) = V^\top A V$ and $\widetilde{G}(V) = V^\top \widetilde{A} V$. Note, then, that the local optimality condition is equivalent to 
\begin{align}
	AVE(V) - \alpha \widetilde{A} V \widetilde{E}(V)
	= BVE(V) G(V) E(V) - \alpha \widetilde{B} V \widetilde{E}(V) \widetilde{G}(V) \widetilde{E}(V).\label{eqn:8-it}
\end{align}

In Appendix \ref{apdx:8-iter}, we discuss how Equation \eqref{eqn:8-it} may lead to a gradient-free algorithm that involves solving an asymmetric algebraic Ricatti equation. 



So far, we have discussed the population version, which relies on the distributions of $x,~\widetilde{x},~y,$ and $\widetilde{y}$ that are unknown in practice. In real applications, we observe finite samples $(x_i,y_i)_{i=1}^n$ as foreground data and  $(\widetilde{x}_j,\widetilde{y}_j)_{j=1}^m$ as background data. We denote $X\in\RR^{n\times p}$, $\widetilde{X}\in\RR^{m\times p}$ where each row represents a sample; similarly, each entry of $Y\in\RR^n$ and $\widetilde{Y}\in\RR^m$ represents a response value. In this case, we can replace the expectation by the sample mean to get estimates of $\Sigma_x,\ \Sigma_{\widetilde{x}},\ \Sigma_{xx},\ \text{ and } \Sigma_{\widetilde{x}\widetilde{x}}$ and have the corresponding plug-in estimates for $A,\ B,\ \widetilde{A},\ \text{and } \widetilde{B}$. Then, optimization reduces to a manifold optimization problem \citep{absil_optimization_2009}. The estimates of $\Sigma_x$ and $\Sigma_{\widetilde{x}}$ deserve further discussion. As shown by \citet{li1991sliced} and \citet{cai2020online} among others, for continuous response $y$, the observed support of response $y$ can be discretized into \emph{slices} $I_{h}=(q_{h-1},q_h]$, for $h=1,\cdots,H$. An estimate of $\Sigma_x$ is given by $\sum_{h=1}^H m_hm_h^\top$ where $m_h = \frac{1}{np_h}\sum_{y_i\in I_h} x_i$ with $p_h =\frac{1}{n}\sum_{i=1}^n I(y_i\in I_h)$. On the other hand, if $y$ and $\widetilde{y}$ are categorical, the slices are naturally chosen as all possible values of $y$ and $\widetilde{y}$. Combining these pieces, we present our empirical version of the CIR algorithm in Algorithm \ref{alg:CIR}.

It is worth noting that our optimization of $f$ as a function of $V\in \stf{(p, d)}$ cannot be considered as an optimization problem in $\RR^{p\times d}$ with orthogonality constraints 
$V^\top V= \id_d$~\citep{edelman1998geometry, boumal2019global}. Because the term $(V^\top BV)^{-1}$ in $f$ is not well defined unless $V $ is full rank, our loss function $f$ cannot be extended to the full Euclidean space $\RR^{p\times d}$. We consider it as an optimization problem \emph{intrinsically defined} on $\stf{(p, d)}$ as laid out by \citet{absil_optimization_2009}. This key property excludes some commonly used optimizers on manifolds, and we will discuss more details in the next section.

\begin{algorithm}[tb]
	\caption{CIR}
	\label{alg:CIR}
	\begin{algorithmic}
		\STATE {\bfseries Input:} Foreground data $(X, Y) \in \mathbb{R}^{n \times p}\times \RR^{n}$,\ Background data $(\widetilde{X}, \widetilde{Y}) \in \mathbb{R}^{m \times p}\times \RR^{m}$, $\alpha > 0$, $d\in\ZZ_+$.\medskip
		\STATE $x_i = x_i-\frac{1}{n}\sum_{i=1}^nx_i$;~$\widetilde{x}_j = \widetilde{x}_j-\frac{1}{m}\sum_{j=1}^m\widetilde{x}_j$ \medskip
		\STATE  $\Sigma_{xx} = \frac{1}{n}\sum_{i=1}^n x_ix_i^\top$;~$\Sigma_{\widetilde{x}\widetilde{x}} = \frac{1}{m}\sum_{j=1}^m \widetilde{x}_j\widetilde{x}_j^\top$.\medskip
		\FOR{$h = 1, \dots, H$}
		\STATE Calculate slice proportions $p_h = \frac{1}{n} \sum_{i = 1}^n I(y_i \in I_h)$.
		\STATE  Calculate slice mean $m_h = \frac{1}{np_h} \sum_{y_i \in I_h} x_i$.
		\ENDFOR
		\STATE  $\Sigma_x = \sum_{h = 1}^{H} m_h m_h^\top$.\medskip
		
		\FOR{$\widetilde{h} = 1, \dots, \widetilde{H}$}
		\STATE Calculate slice proportions $p_{\widetilde{h}} = \frac{1}{m} \sum_{j = 1}^m I(\widetilde{y}_j \in I_{\widetilde{h}})$.
		\STATE Calculate slice mean $m_{\widetilde{h}} = \frac{1}{mp_{\widetilde{h}}} \sum_{\widetilde{y}_j \in I_{\widetilde{h}}} \tilde{x}_j$.
		\ENDFOR
		\STATE $\Sigma_{\widetilde{x}} = \sum_{\widetilde{h} = 1}^{\widetilde{H}} m_{\widetilde{h}}m_{\widetilde{h}}^\top$.\medskip
		
		\STATE Compute $A = \Sigma_{xx} \Sigma_x \Sigma_{xx}$,\ $B=\Sigma_{xx}^2$,\\ \ \ \ \ \ \ \ \ \ \ \ \ \ \ \ \ $\tilde{A} = \Sigma_{\widetilde{x}\widetilde{x}}\Sigma_{\widetilde{x}}\Sigma_{\widetilde{x}\widetilde{x}}$,\ $\widetilde{B}=\Sigma_{\widetilde{x}\widetilde{x}}^2$.
		\STATE Find $V^* = \arg\min_{V \in \stf{(p, d)}} f(V;\ A,\ B,\ \widetilde{A},\ \widetilde{B},\ \alpha)$ for $f$ defined in \eqref{eqn:cir_loss}.\medskip
		\STATE \bf{Return} $V^*$.
	\end{algorithmic}
\end{algorithm}

\section{Theory}
\label{sec:theory}
In this section, we discuss two concrete optimization algorithms for the last step in Algorithm \ref{alg:CIR} to find $V^*$ and show their convergence. The optimization problem outlined in Equation \eqref{eqn:cir_loss} does not follow the classic Li-Duan theorem for regression-based dimension reduction (See e.g., \citet{cook2009regression}, Prop 8.1), due to its nonconvex nature. The convergence of the algorithm is discussed in detail below. 

The first algorithm we consider is the scaled gradient projection method (SGPM) specifically designed for optimization on the Stiefel manifold~\citep{oviedo2019scaled}. We first define an analog to Lagrangian multiplier $\mathcal{L}(V,\Lambda )\coloneqq f(V)-\frac{1}{2}\<\Lambda,V^\top V-\id_d\>$, then the SGPM algorithm is summarized in Algorithm \ref{alg:SGPM}, where $\pi(X) = \arg\min_{Q \in \stf{(p, d)}} ||X - Q||_F$ is the orthogonal projection to the Stiefel manifold. Note for Algorithm \ref{alg:SGPM} that the update for $\mu$ and $C_{k+1}$ is intricate; see \citet{oviedo2019scaled} for more details.
		
		
		

\begin{algorithm}[tb]
	\caption{SGPM \citep{oviedo2019scaled})}
	\label{alg:SGPM}
	\begin{algorithmic}
		\STATE {\bfseries Input:} $V_0 \in \stf{(p, d)},\ \eta \in [0, 1],\ \mu, \rho_1, \epsilon, \delta \in (0, 1)$.
		\STATE Set $Q_0 = 1, C_0 = f(V_0)$, and $k = 0$
		\WHILE{$\left\| \nabla_V \mathcal{L}(V_k) \right\| > \epsilon$}
		\STATE Set $A = \nabla f(V_k) V_k^{\top} - V_k \nabla f(V_k)^\top$
		\STATE Set $D_{\mu, \tau_k} = (I_p + \mu \tau_k A)^{-1}$
		\STATE Let $Y(\tau) := V_k - \tau (I_p - \mu \tau A)^{-1} \nabla f(V_k)$
		\STATE Pick $\tau_k>0$ so that $f(Y(\tau_k)) > C_k + \rho_1 \tau_k Df(X_k) [\dot{Y}(0)]$
		\STATE Update $V_{k + 1} = \pi\left(V_k - \tau D_{\mu, \tau} \nabla f(V_k) \right)$
		
		\STATE Update $\mu$ and $C_{k+1}$
		
		
		
		\STATE Set $k = k + 1$
		\ENDWHILE
		\STATE $V^* = V_k$
	\end{algorithmic}
\end{algorithm}

To study the convergence of Algorithm \ref{alg:SGPM}, we need to study the Karush-Kuhn-Tucker (KKT) conditions for CIR:
\begin{definition}[\citet{oviedo2019scaled}]\label{lem:kkt}
	The KKT conditions are given by
	\begin{align*}
		D_V\mathcal{L}(V,\Lambda) &= \nabla f- V\Lambda = 0\\
		D_\Lambda\mathcal{L}(V,\Lambda) & = V^\top V-\id_d.
	\end{align*}
\end{definition}
Now we can state the convergence theorem of Algorithm \ref{alg:SGPM}:
\begin{theorem}\label{thm:sgpm}
	Let $\{V_k\}_{k=1}^\infty$ be an infinite sequence generated by Algorithm \ref{alg:SGPM}, then any accumulation point $V_*$ of $\{V_k\}_{k=1}^\infty$ satisfies the KKT conditions in Lemma \ref{lem:kkt}, and $\lim_{k\to\infty} \|D_V\mathcal{L}(V_k)\|=0$.
\end{theorem}

\begin{algorithm}[tb]
	\caption{ ALS~\citep{absil_optimization_2009}}
	\label{alg:LS}
	\begin{algorithmic}
		\STATE {\bfseries Input:} $V_0 \in \stf{(p, d)}$, retraction $R$ from $T\stf{(p, d)}$ to $\stf{(p, d)}$; scalars $\overline{\alpha} > 0$, $c, \beta, \sigma \in (0, 1)$.
		\FOR{$k = 0, 1, \dots$}
		\STATE $\eta_k =-\grad f(V_k)$
		\STATE Select $V_{k + 1}$ so that $f(V_k) - f(V_{k + 1}) \geq c (f(V_k) - f(R_{V_k}(t_k^A\eta_k)))$
		\ENDFOR
	\end{algorithmic}
\end{algorithm}

Although Algorithm \ref{alg:SGPM} is guaranteed to converge, there are two drawbacks. First, the accumulation point $V_*$ is only guaranteed to satisfy the KKT conditions, but not necessarily be a critical point. Second, we do not know how fast $V_k$ will converge to $V_*$. Next, we introduce an accelerated line search (ALS) algorithm as an alternative to SGPM that converges to a critical point with a known convergence rate. ALS is summarized by Algorithm \ref{alg:LS}, where $t_k^A$ is the step size, called the Armijo step size for given $\overline{\alpha}, \beta, \sigma, \eta_k$ and $R$ is a retraction to $\stf(p,d)$, see \citet{absil_optimization_2009} for more details.

Algorithm \ref{alg:LS} can be shown to have linear convergence to critical points if the hyperparameters are chosen properly. For other choices of $\eta_k$, see Appendix~\ref{sec:gradient_based} for more details. The following adaptation of Theorem 4.5.6 in \citet{absil_optimization_2009} indicates linear convergence to stationary points.

\begin{theorem}\label{thm:ls}
	Let $\{V_k\}_{k=1}^\infty$ be an infinite sequence generated by Algorithm \ref{alg:LS} with $\eta_k=-\grad~f(V_k)$ converging to an accumulation point $V_*$ of $\{V_k\}_{k=1}^\infty$, then $V_*$ is a critical point of $f$, and $\lim_{k\to\infty} \|\grad{f}(V_k)\|=0$. 
	
	Furthermore, assuming $V_*$ is a local minimizer of $f$ with $0<\lambda_l\coloneqq\min\eig(\Hess(f)(V_*))$ and $\lambda_u\coloneqq\max\eig(\Hess(f)(V_*))$, then, for any $r\in(r_*,1)$ where $$r_*\coloneqq 1-\min\left(2\sigma\bar{\alpha}\lambda_l,4\sigma(1-\sigma)\beta\frac{\lambda_l}{\lambda_u}\right),$$ there exists an integer $K\neq 0$ such that $$f(V_{k+1})-f(V_*)\leq (r+(1-r)(1-c))(f(V_k)-f(V_*)),$$ for all $k\geq K$, where $\bar{\alpha},\beta,\sigma,c$ are the hyperparameters in Algorithm \ref{alg:LS}.
\end{theorem}

The difference between Algorithm \ref{alg:SGPM} and \ref{alg:LS} deserves further comment. While we empirically observe that Algorithm \ref{alg:SGPM} often converges faster, Algorithm \ref{alg:LS} has theoretical properties which allow for a proof of linear convergence in terms of an upper bound on the number of iterations. In practice, for smaller datasets we suggest running Algorithm \ref{alg:LS}, while for larger datasets we recommend using Algorithm \ref{alg:SGPM} for efficiency. 

The computational complexity of CIR for both the SGPM-based optimization and the ALS-based optimization is compared to various competitors in the table below. Here, we assume that $1 \leq d < p < m, n$, where the background and foreground data have $m$ and $n$ samples, respectively. Specifically, we assume that $\epsilon$ denotes the stopping error such that $f(V_{k}) - f(V_*) \leq \epsilon$. It is noteworthy that a $p$-dimensional singular value decomposition can be achieved within $\mathcal{O}(p^3)$. We present the comparison in the table below. 
\begin{table}[!h]
	\caption{Computational Time-Complexity of CIR and Competitors}
	\begin{tabular}{c| c} 
		\hline
		Algorithm & Theoretical Algorithmic Complexity \\ [0.5ex] 
		\hline
		CIR, SGPM-based & $\mathcal{O}(  (m + n)p^2)$ \\ 
		\hline
		CIR, ALS-based & $\mathcal{O}(-\log(\epsilon) p^{3} + (m + n)p^2)$ \\
		\hline
		LDA & $\mathcal{O}(np^2)$ \\
		\hline
		PCA & $\mathcal{O}(np^2)$ \\
		\hline
		CPCA & $\mathcal{O}((m+n)p^2)$ \\
		\hline
		SIR & $\mathcal{O}(np^{2})$ \\
		\hline
	\end{tabular}
\end{table}


\section{Application}
\label{sec:appl}
When applying CIR, several hyperparameters must be tuned, such as the weight $\alpha$, the reduced dimension $d$, and the slices $I_h$ and $I_{\widetilde{h}}$ for estimation of $\Sigma_x$ and $\Sigma_{\widetilde{x}}$. In some cases, it may also be necessary to determine the definition of the foreground and background groups and to assign background labels $\widetilde{Y}$.

The value of $\alpha \geq 0$ can be determined by cross-validation. Our numerical experiments show that CIR is robust to the choice of $\alpha$; that is, the performance of the method changes continuously with $\alpha$. Tables supporting this observation are provided in \ref{apdx:mouse} and \ref{apdx:sc}.

Additionally, the choice of reduced dimension $d$ may depend on the goal of the investigator. If visualization is considered important, $d = 2$ is appropriate. If the goal is prediction, the elbow point of the $d$ versus prediction error plot may suggest an optimal $d$. However, as with other DR methods, determining the optimal value of $d$ is still a topic of ongoing research~\cite{camastra2016intrinsic, campadelli2015intrinsic}.

The definition of foreground data $X$ and $Y$ should be the data and the target variable of interest, while the choices of background data $\widetilde{X}$ and $\widetilde{Y}$ may not be as straightforward. These data are intended to represent ``noise" that is ``subtracted" from the foreground data. For example, in the biomedical context, if the population of interest is a group of sick patients, the background dataset may include observations of healthy individuals. In other contexts, however, it may be appropriate to use $\widetilde{X} = X$. In this case, the choice of background label $\widetilde{Y}$ may be unclear. If another outcome variable was collected, it could be used as $\widetilde{Y}$; otherwise, randomly selected values in the support of $Y$ could be used to represent ``noise".

The estimates for $\Sigma_x$ and $\Sigma_{\widetilde{x}}$ are partly determined by whether $y$ and $\widetilde{y}$ are categorical or continuous. If these variables are categorical, then each value of $y$ (or $\widetilde{y}$) can be considered as a separate slice, resulting in $|\text{supp}(Y)|$ (or $|\text{supp}(\widetilde{Y})|$) total slices. On the other hand, if these variables are continuous, slices can be taken to represent an equally spaced partition of the range of $Y$ (or $\widetilde{Y}$), with the number of slices being tunable hyperparameters.

\subsection{Mouse Protein}
The first dataset we considered was collected for the purpose of identifying proteins critical to learning in a mouse model of Down syndrome \cite{higuera2015self}. The data contain 1095 observations of expression levels of 77 different proteins, along with genotype~(t=Ts65Dn, c=control), behavior~(CS=context-shock, SC=shock-context), and treatment~(m=memantine, s=saline). The behavior of CS corresponds to the scenario in which the mouse was first placed in a new cage and permitted to explore for a few minutes before being exposed to a brief electric shock; conversely, SC corresponds to mice immediately given an electric shock upon being placed in a new cage, and then being permitted to explore. Of the data, 543 samples contain at least one missing value. Taking into account the relatively large sample size, we consider only the 552 observations with complete data. We do not perform any normalization or any other type of preprocessing to the raw data prior to analysis.

\begin{figure}[!h]
	\includegraphics[width = \columnwidth]{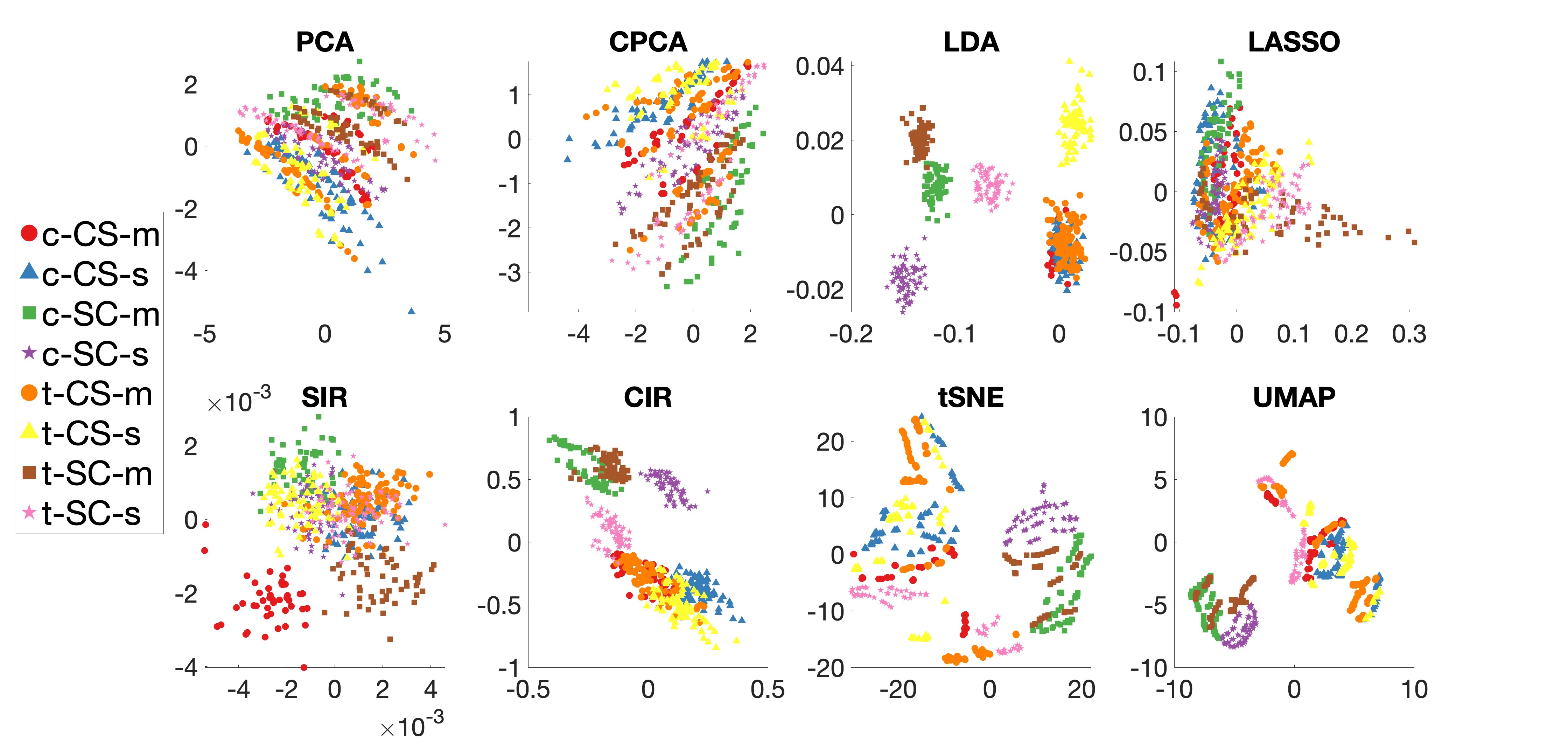}
	\caption{\label{fig:Mouse_vis}2-d representation of mouse protein data}
\end{figure}

In this example, $X\in\RR^{552\times 77}$ represents the expression of 77 proteins of all mice without a missing value, while $y_i\in\{0,1,\cdots,7\}$ represents the combination of 3 binary variables: genotype, treatment, and behavior. For example, $y_i=1$ means that the $i$-th mouse received memantine, was exposed to context-shock, and has genotype Ts65Dn. To visualize the data, we apply unsupervised DR algorithms PCA, tSNE and UMAP to $X$ and supervised DR methods LDA, LASSO and SIR to $(X,Y)$, with $d=2$ for all algorithms. The 2-dimensional representation is given in Figure \ref{fig:Mouse_vis}, where each color represents a class of mice among 8 total classes. 

PCA, LASSO, SIR, tSNE, and UMAP fail to distinguish between classes, whereas LDA successfully separates 5 classes but with 3 classes (c-CS-m, t-CS-m, t-CS-s) mixed together. Now we take advantage of the background data. We let $\widetilde{X}$ be the protein expression from mice with genotype = control, which coincides with the background group used in previous studies of this application \citep{abid2018exploring,li2020probabilistic}. We set $\widetilde{Y}$ as the binary variable representing behavior and apply CPCA to $(X,\widetilde{X})$ and CIR to $(X,Y,\widetilde{X},\widetilde{Y})$ with $d=2$ as well. The 2-dimensional representations are shown in Figure \ref{fig:Mouse_vis}, which indicates that CIR outperforms all other competitors. In particular, the three classes that were not separated in LDA are less mixed in CIR.

\begin{figure}[!h]
	\centering
	\includegraphics[width = 0.75\textwidth]{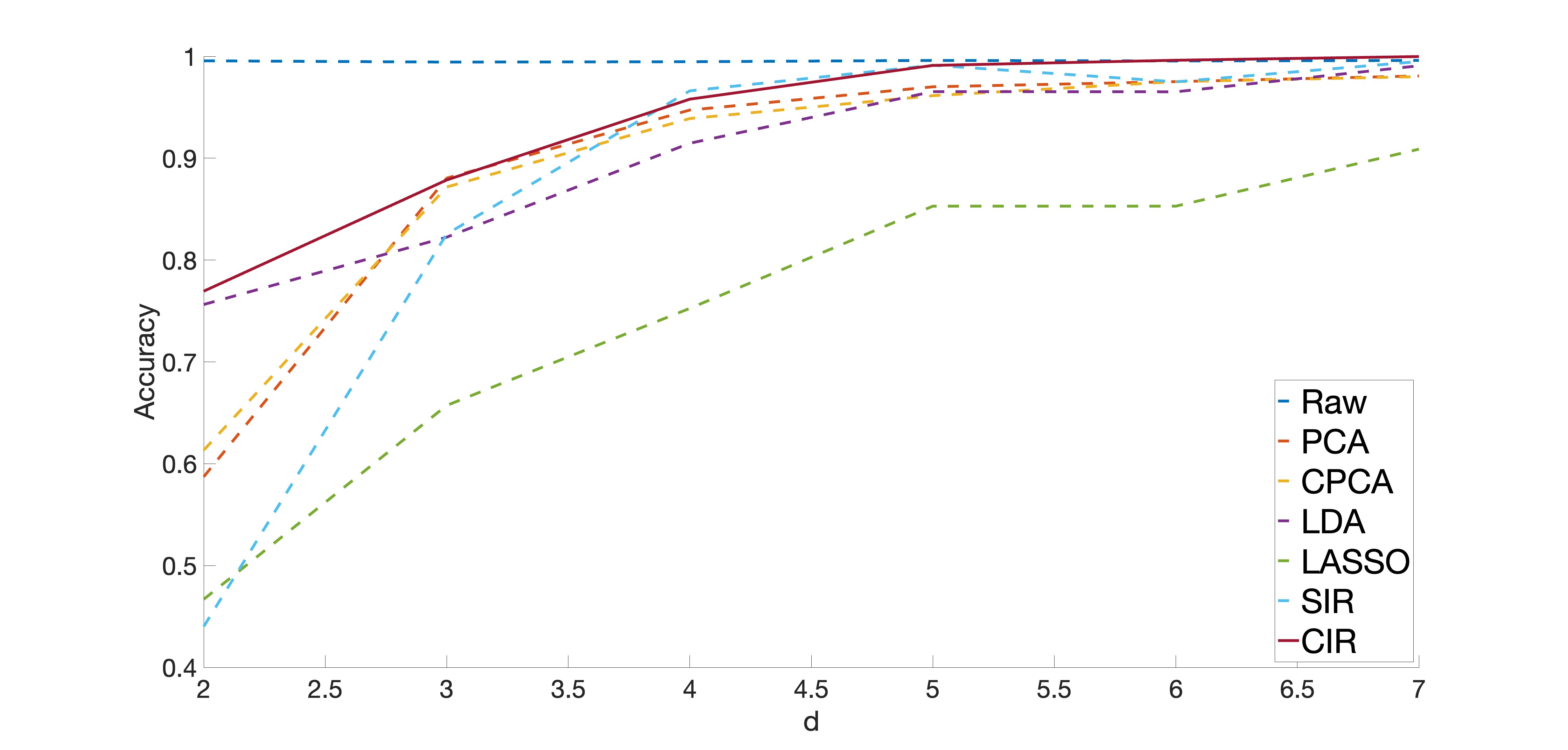}
	\caption{\label{fig:mouse_knn}Classification accuracy by KNN for mouse protein data}
\end{figure}

Next, we show the classification accuracy based on $XV$, the dimension-reduced data. Here, we vary $d$ from $2$ to $7$ because for higher $d$, the accuracy is close to $1$. The mean prediction accuracy of KNN, the best classifier for this example, over 10 replicates versus the reduced dimension $d$ is shown in Figure \ref{fig:mouse_knn}, which clearly indicates that CIR outperforms all competitors especially when $d$ is small. We present the accuracy of other classifiers and their standard deviations in Appendix \ref{apdx:mouse}. 

\subsection{Single Cell RNA Sequencing}

The second dataset we considered is from a study of single-cell RNA sequencing used to classify cells into cell types based on their transcriptional profile \cite{alquicira2019scpred}. The data include $3500$ observations of expression levels of $32838$ different genes, along with labels of the cell as one of 9 different cell types, namely CD8 T cell, CD4 T cell, classical monocyte, B cell, NK cell, plasmacytoid dendritic cell, non-classical monocyte, classic dendritic cell, and plasma cell. We select the top 100 most variable genes for our analysis to be consistent with previous analyses of these data \cite{zheng2017massively,abid2018exploring}. In this example, $X \in \mathbb{R}^{3500\times 100}$ represents the expression of $100$ genes, while $y_i \in \{0, 1, \dots, 8\}$ represents the cell type. For example, $y_i = 1$ means that the $i$-th cell is a CD4 T cell.

\begin{figure}[!h]
	\includegraphics[width = \columnwidth]{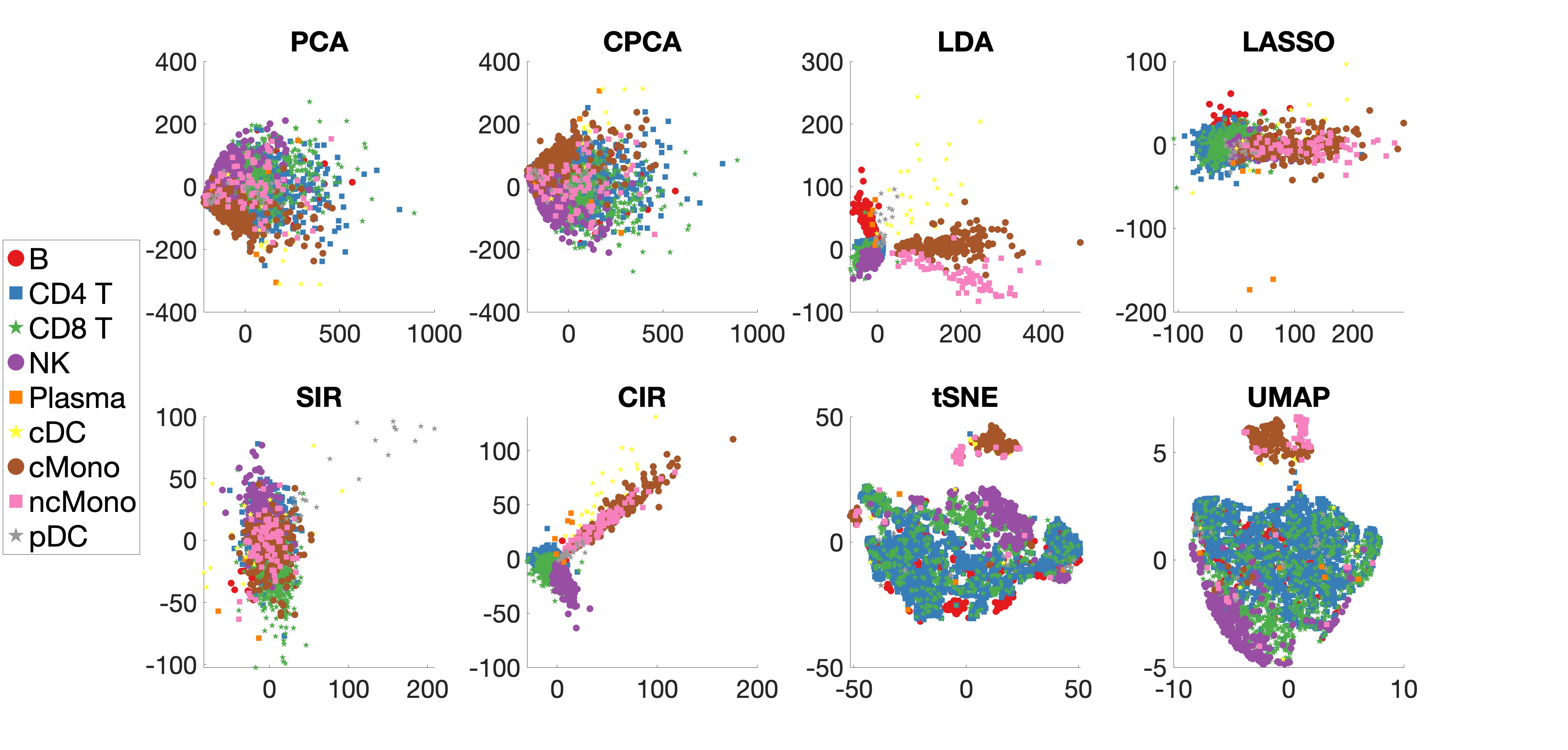}
	\caption{\label{fig:sc_vis}2-d representation of single-cell RNA sequencing data}
\end{figure}

To visualize the data, we apply unsupervised DR algorithms PCA, tSNE, and UMAP to $X$ and supervised DR methods LDA, LASSO, and SIR to $(X, Y)$, for $d = 2$ for all algorithms. In this case, there is no obvious choice of background data. So, we use $\widetilde{X} = X$ and randomly draw independent and identically distributed samples $\widetilde{Y} \sim \text{uniform}\{0,\cdots,8\}$ in order to apply CPCA and CIR. 
The 2-dimensional representation is given in Figure \ref{fig:sc_vis}.

\begin{figure}[!h]
	\centering
	\includegraphics[width = 0.75\textwidth]{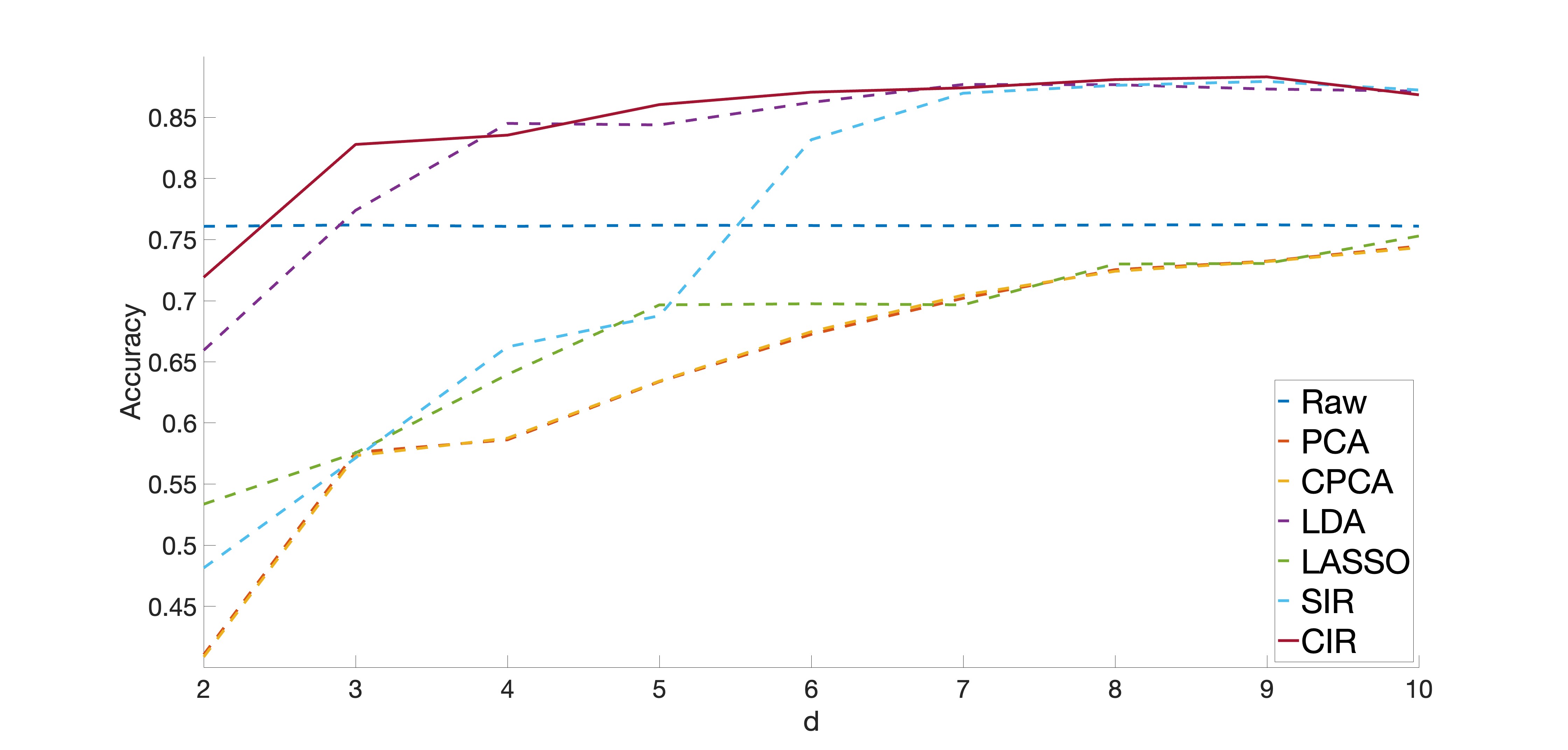}
	\caption{\label{fig:sc_accuracy}Classification accuracy by KNN for single-cell RNA sequencing data}
\end{figure}

For each $d=2,\cdots,10$, we compare the accuracy of a KNN classifier based on dimension-reduced data among various methods, with the raw data as the baseline. We repeat this process $10$ times to reduce the impact of random split in cross-validation. The prediction accuracy versus reduced dimension $d$ is shown in Figure \ref{fig:sc_accuracy}, where CIR has the best overall performance especially when $d=2,3$. We show the accuracy of other classifiers and their standard deviations in Appendix \ref{apdx:sc}.

The improved performance of CIR over SIR deserves further comment. While the background data and labels $(\widetilde{X}, \widetilde{Y})$ used in CIR do not add new information beyond what SIR used, because the background label is chosen randomly, we attribute the improved performance to CIR ``denoising" the foreground data.

\subsection{Plasma Retinol}

The third dataset we consider is the plasma retinol dataset~\cite{nierenberg1989determinants}. The dataset contains 315 observations of 14 variables, including age, sex, smoking status, BMI, vitamin use, calories, fat, fiber, cholesterol, dietary beta-carotene, dietary retinol consumed per day, number of alcoholic drinks consumed per week, and levels of plasma beta-carotene and plasma retinol.

In this example, $X \in \mathbb{R}^{315\times 12}$ represents measurements of the first 12 variables listed for all subjects, while $y_i$ represents the measurement of plasma beta-carotene, a variable of particular interest to scientists~\cite{nierenberg1989determinants}. In contrast to the previous two examples, note that here $y_i$ is continuous, not categorical.

We apply unsupervised DR algorithms PCA, tSNE, and UMAP to $X$ and supervised DR algorithms LDA, LASSO, and SIR to $(X, Y)$ for $d = 1, \dots, 8$. Similarly to the single-cell RNA sequencing application, we let $\widetilde{X} = X$ because there is no natural choice of background data. For the background label, we set $\widetilde{Y}$ as the continuous variable representing the level of plasma retinol, which shares certain information with $y_i$, and apply CPCA to $(X, \widetilde{X})$ and CIR to $(X, Y, \widetilde{X}, \widetilde{Y})$ for $d = 1, \dots, 8$. We skip the visualization step in this case due to the poor visibility in terms of $y_i$.

\begin{figure}[!h]
	\centering
	\includegraphics[width = 0.75\textwidth]{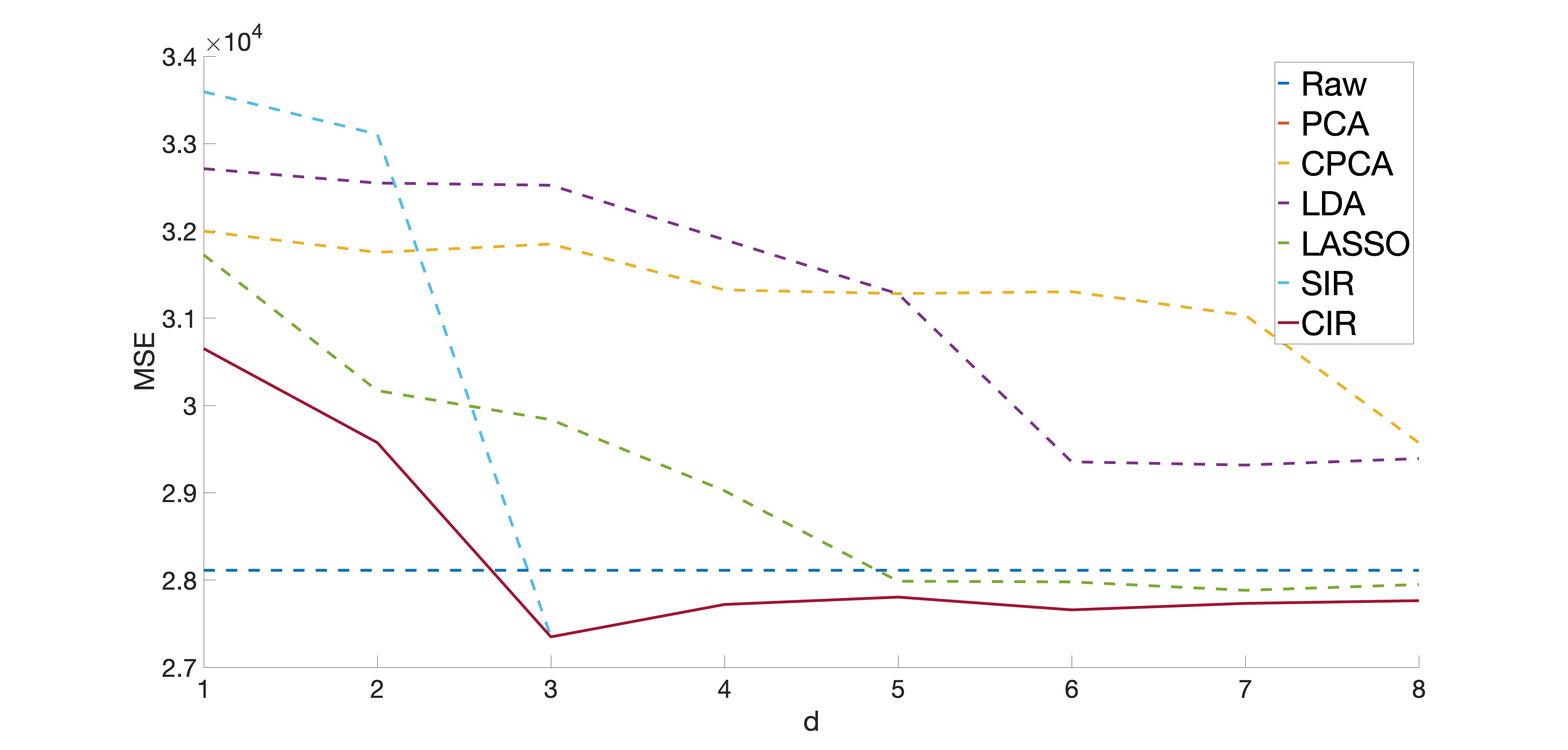}
	\caption{\label{fig:plasma_mse}MSE of linear regression for plasma retinol data}
\end{figure}

After trying a few regression methods, namely linear regression~\cite{freedman2009statistical}, regression trees~\cite{breiman2017classification}, Gaussian process regression~\cite{chen2020multivariate}, and neural networks~\cite{hopfield1982neural}, we present the prediction mean squared error (MSE) for the best method for this dataset, linear regression. That is, for each $d$ and the output $V$ from each DR algorithm, we fit a linear regression model to $(XV, Y)$. We also compare to a linear regression model based on raw data $(X, Y)$ as the baseline. Figure \ref{fig:plasma_mse} demonstrates that CIR outperforms all other competitors, but matches SIR when $d\geq 3$. We display the MSE of other regression methods and their standard deviations in Appendix \ref{apdx:plasma}.

Note that because $Y$ and $\widetilde{Y}$ are continuous, the number of slices to estimate $\Sigma_x$ and $\Sigma_{\widetilde{x}}$ needs to be carefully chosen and adjusted to ensure optimal performance. 
We use cross-validation to select 4 equally spaced partitions for the support of $Y$ and $\widetilde{Y}$.
In the three applications presented above, CIR demonstrates superior overall performance over its supervised, unsupervised, contrastive, and non-contrastive competitors, especially in low dimension, i.e., $d=2,3$, which are the most crucial dimensions for visualization purposes.

\section{Discussion and Future Work}
\label{sec:disc}
In this work, we propose the CIR model and the associated optimization algorithm for supervised dimension reduction for datasets that are split into foreground and background groups. We provide theoretical guarantee of the convergence of the CIR algorithm under mild conditions. We have shown that our CIR model outperforms competitors in multiple biomedical datasets, including mouse protein expression data, single-cell RNA sequencing data, and plasma retinol data. However, there are several important future directions that remain unaddressed in this paper, as outlined below.

\textbf{Multi-treatment.} It is natural to consider how our model can be extended to studies with multiple treatments. For example, in medical treatment, there might be more than one treatment for patients with certain disease. In \citet{vogelstein2021supervised}, it has been shown that the number of treatment groups puts a hard constraint on the target dimension. 
It is interesting to generalize from a single-treatment structure to a multi-treatment structure (e.g., \citet{luo2022nonparametric}), where the loss function needs more sophisticated design.

Another direction in multi-group scenario is to combine multiple CIR models trained on different pairs of bi-groups. As pointed out by \citet{wolpert1992stacked}, the generalization error in the constrative regression model stacking needs to be controlled, and one possible way is to follow divergence mixing as proposed by \citet{hinton2002training}, with a careful normalization. The major difficulty in training such stacking model is how to devise a sequential optimization for model training. 

\textbf{Consistency and sufficient dimension reduction.}
he consistency of the proposed CIR model remains open. Theorems \ref{thm:sgpm} and \ref{thm:ls} ensure that the resulting solution must be a stationary point, but we did not discuss whether these stationary points are  consistent estimates. The consistency of the estimates is also affected by the choice of $\alpha$, because $\alpha=0$ will render this CIR into a classical SIR for the foreground group. This consistency problem also has practical importance, as it explicitly expresses the trade-off between the expressive contrastiveness and the emphasis on the effective lower-dimensional structures. The group information and statistical sufficiency compete against each other, as we observed in the experiments, thus a range of $\alpha$ that balances between these two factors are of interest and might be answered by the consistency result.

Furthermore, SIR has the drawback of missing the totality central subspace when the symmetry assumption in $x$ is lost~\citep{li2018sufficient}. \citet{cook1991sliced} proposed the sliced average variance estimator~(SAVE) estimator for addressing this problem, which raises the natural question of how to generalize this high-moment SDR method to the contrastive setting.

\textbf{Loss function.} Our loss function \eqref{eqn:cir_loss} is nonstandard, which raises many questions. For example, the relation between number of local minima and $A,B,\widetilde{A},\widetilde{B},\alpha$ remains open. Moreover, although $f$ cannot be continuously extended to the full Euclidean space $\RR^{p\times d}$, if we restrict the domain to be a submanifold of $\stf(p,d)$, $f$ might be extended to the convex hull of the submanifold. This extension will enable us to apply some other efficient optimization algorithms with strong theoretical guarantee~\cite{boumal2019global}. Furthermore, Appendix \ref{apdx:8-iter} raises the question of the validity of a fixed-point algorithm based on Ricatti equations that may lead to a more efficient algorithm to minimize $f$ without involving the gradient.

\bibliographystyle{apalike}
\bibliography{ref.bib}

\newpage

\appendix
\onecolumn
\section{Proof to Lemma \ref{lem:grad}}
\label{proof_lem:grad}

\begin{align*}
	- \frac{\partial f}{\partial V} &= 2AV(V^\top BV)^{-1}-2BV (V^\top BV)^{-1} V^\top AV (V^\top BV)^{-1}\\
	& ~~~-\alpha\left\{2\widetilde{A}V(V^\top \widetilde{B}V)^{-1}-2\widetilde{B}V (V^\top \widetilde{B}V)^{-1} V^\top \widetilde{A}V (V^\top \widetilde{B}V)^{-1}\right\}.
\end{align*}
Recall that the projection to tangent space of the Stiefel manifold $\stf(p,d)$ at $V$ is given by 
$$\proj_V(Z) = Z-V\sym(V^\top Z),~\forall Z\in T_V\stf(p,d),$$
where $\sym(X)\cdots \frac{X+X^\top}{2}$ is the symmetrizer. Then observe that the following equations involving the pair $A,B$ and the pair $\widetilde{A},\widetilde{B}$ have to satisfy the following equations: 
	\begin{align*}
		V^\top \left(2AV(V^\top BV)^{-1}-2BV (V^\top BV)^{-1} V^\top AV (V^\top BV)^{-1}\right)& = 0\\
		V^\top \left(2\tdA V(V^\top \tdB V)^{-1}-2\tdB V (V^\top \tdB V)^{-1} V^\top \tdA V (V^\top \tdB V)^{-1}\right)& = 0.
	\end{align*}
	
That is, $V^\top \frac{\partial f}{\partial V}=0$. As a result, the gradient of $f$ is given by
\begin{align*}
	-\grad f(V) &= \proj_V(\frac{\partial f}{\partial V}) =\frac{\partial f}{\partial V}- V\sym(V^\top \frac{\partial f}{\partial V}) = -\frac{\partial f}{\partial V}. 
\end{align*}

\section{Proof of Theorem \ref{thm:critical}}\label{proof_thm:critical}

If $V$ is a local minimizer (i.e., a stationary point for the optimization problem \eqref{eqn:cir_loss}), then $\grad f(V)=0$, from Lemma \ref{lem:grad} we have
\begin{align*}
	AVE(V)-\alpha \widetilde{A}V\widetilde{E}(V) = BVF(V)-\alpha \widetilde{B}V\widetilde{F}(V),
\end{align*}
where $E(V) = (V^\top BV)^{-1}$, $\tdE(V) = (V^\top \tdB V)^{-1}$, $F(V) = (V^\top BV)^{-1} V^\top AV (V^\top BV)^{-1}$ and $\tdF(V) = (V^\top \tdB V)^{-1} V^\top \tdA V (V^\top \tdB V)^{-1}$.


\section{Proof of Theorem \ref{thm:sgpm}}
By Theorem 1 and Corollary 1 in \citet{oviedo2019scaled}, it suffices to show $f$ is continuously differentiable, which is a directly corollary of Lemma \ref{lem:grad}.

\section{Options for $\eta_k$}\label{sec:gradient_based}
To introduce other options for $\eta_k$, we need the following definition. 

\begin{definition} (Gradient-related sequence, see \citet{absil_optimization_2009} (p.62, Definition 4.2.1)).
	Given a function $f$ on a Riemannian manifold $M$, a sequence
	in tangent space $\{\eta_{k}\},\eta_{k}\in T_{V_{k}}M$, where $V_{k}$
	are defined through the iterative formula 
	$V_{k+1}  =R_{V_{k}}(t_{k}\eta_{k}),$
	and $R_{x_{k}}$ can be any retraction (e.g., global retraction
	mapping $\text{Retr}_{V}:T_{V}M\rightarrow M,\xi\mapsto(V+\xi)(I_{d}+\xi^{\top}\xi)^{-1/2}$
	on $St(n,p)$), is called \textbf{gradient-related} if, for any subsequence
	of $\{V_{k}\}_{k\in K\subset\{1,2,\cdots,n\}}$ that converges to
	a non-critical point of $f$, the corresponding subsequence $\{\eta_{k}\}_{k\in K}$
	is bounded and satisfies 
	\begin{align*}
		\lim\sup_{k\rightarrow\infty,k\in K}\left\langle \grad f(V_{k}),\eta_{k}\right\rangle _{M} & <0.
	\end{align*}
\end{definition}
This means that the cosine of gradient and update $\eta_{k}$ needs
to form an acute angle for only critical points. Note that a naive Newton step is not necessarily
gradient-related (see p.122 in \citet{absil_optimization_2009}). 
In particular, $\eta_k=-\grad f(V_k)$ results in a gradient-related sequence, and is suggested by \citet{absil_optimization_2009} as a natural choice.

\section{Proof of Theorem \ref{thm:ls}}
The first assertion regarding consistency is from Theorem 4.3.1 in \cite{absil_optimization_2009}, which requires our loss function $f$ to be continuously differentiable, a direct corollary of Lemma \ref{lem:grad}. 

By the compactness of $\stf(p,d)$, the level set $\mathcal{L}\coloneqq \{V\in \stf(p,d):~f(V)\leq f(V_0)\}$ is compact for any $V_0\in \stf(p,d)$, the second assertion follows Corollary 4.3.2 in \cite{absil_optimization_2009}.

The third assertion regarding convergence rate involves second-order conditions, i.e., the Hessian of $f$.  Let $D^2f$ be the Hessian computed in Euclidean coordinates, that is, $(D^2f|_V)_{ij,kl}\coloneqq \frac{\partial f}{\partial V_{ij}\partial V_{kl}}$, then for tangent vectors $\Omega_1,\Omega_2\in T_{V}\stf(p,d)$, the Hessian is given by \citet{absil_optimization_2009}
\begin{align*}
	\Hess(f)(\Omega_1,\Omega_2)&=\underbrace{D^2f|_V(\Omega_1,\Omega_2)}_{\circled{1}
	}+\underbrace{\frac{1}{2}\tr\left((\grad f(V)^\top \Omega_1 V^\top+V^\top \Omega_1 \grad f(V)^\top)\Omega_2\right)}_{\circled{2}}\\
	&~~~~\underbrace{-\frac{1}{2}\tr\left((V ^\top \grad f(V)+\grad f(V)^\top V)\Omega_1^\top (I_p-VV^\top)\Omega_2\right)}_{\circled{3}}.
\end{align*}

By the definition of $f$, \circled{1} is $C^\infty$ in the Euclidean sense, so is continuous. By the continuity of $\grad f$, \circled{2} and \circled{3} are also continuous since they are product or summation of continuous functions. Then the convergence rate follows Theorem 4.5.6 in \cite{absil_optimization_2009}.

\section{Fixed-Point Approach to Optimization}\label{apdx:8-iter}

Motivated by the first order optimality condition for the loss function \eqref{eqn:cir_loss}, we seek a fixed-point method as an alternative to a gradient descent-based algorithm. Instead of solving quation \eqref{eqn:8-it} in one algebraic step, we separate the problem into the following 8 equations, which can be solved cyclically. Recall that $E(V) = (V^\top BV)^{-1}$, $\tdE(V) = (V^\top \tdB V)^{-1}$, $G(V) = V^\top A V$, and $\widetilde{G}(V) = V^\top \widetilde{A} V$ and suppress the index of $V_k$, i.e., $V=V_k$ for now for legibility:
\begin{align} 
	AV_{k+1}E(V) - \alpha \widetilde{A} V \widetilde{E}(V) &= BVE(V) G(V) E(V) - \alpha \widetilde{B} V \widetilde{E}(V) \widetilde{G}(V) \widetilde{E}(V)\\
	AVE(V) - \alpha \widetilde{A} V_{k+1} \widetilde{E}(V) &= BVE(V) G(V) E(V) - \alpha \widetilde{B} V \widetilde{E}(V) \widetilde{G}(V) \widetilde{E}(V)\\
	AVE(V) - \alpha \widetilde{A} V \widetilde{E}(V) &= BV_{k + 1} E(V) G(V) E(V) - \alpha \widetilde{B} V \widetilde{E}(V) \widetilde{G}(V) \widetilde{E}(V)\\
	AVE(V) - \alpha \widetilde{A} V \widetilde{E}(V) &= BVE(V) G(V) E(V) - \alpha \widetilde{B} V_{k + 1} \widetilde{E}(V) \widetilde{G}(V) \widetilde{E}(V)\\
	AVE(V_{k + 1}) - \alpha \widetilde{A} V \widetilde{E}(V) &= BVE(V_{k + 1}) G(V) E(V_{k + 1}) - \alpha \widetilde{B} V \widetilde{E}(V) \widetilde{G}(V) \widetilde{E}(V)\\
	AVE(V) - \alpha \widetilde{A} V \widetilde{E}(V_{k + 1}) &= BVE(V) G(V) E(V) - \alpha \widetilde{B} V \widetilde{E}(V_{k + 1}) \widetilde{G}(V) \widetilde{E}(V_{k + 1})\\
	AVE(V) - \alpha \widetilde{A} V \widetilde{E}(V) &= BVE(V) G(V_{k + 1}) E(V) - \alpha \widetilde{B} V \widetilde{E}(V) \widetilde{G}(V) \widetilde{E}(V)\\
	AVE(V) - \alpha \widetilde{A} V \widetilde{E}(V) &= BVE(V) G(V) E(V) - \alpha \widetilde{B} V \widetilde{E}(V) \widetilde{G}(V_{k + 1}) \widetilde{E}(V)
\end{align}

In each of the first four of these equations, $V_{k + 1}$ can change independently, suggesting a convenient corresponding update rule. For the next two equations, we can premultiply by $\left[(BV)^\top (BV)\right]^{-1} (BV)^\top$ (and $\left[(\widetilde{B}V)^\top (\widetilde{B}V)\right]^{-1} (\widetilde{B}V)^\top$, respectively) to obtain the following equation: 
\begin{equation}\label{eqn:Ricatti}
	\left[(BV)^\top (BV)\right]^{-1} (BV)^\top AV E(V_{k+1}) - E(V_{k+1}) G(V) E(V_{k+1}) = \alpha \left[(BV)^\top (BV)\right]^{-1} (BV)^\top H_1,
\end{equation} where $H_1 = \left(\widetilde{A}V\widetilde{E}(V) - \widetilde{B}V\widetilde{E}(V) \widetilde{G}(V) \widetilde{E}(V) \right)$. 

In practice, the cyclic update may not converge to stationary points of the optimization problem \eqref{eqn:cir_loss}. 
However, when the designated cyclic update converges, it can be shown that equation \eqref{eqn:Ricatti} is in the form of an asymmetric algebraic Riccati equation in $E(V_{k + 1})$ \cite{bini2008nonsymmetric}. When we obtain a solution $E^* = E(V_{k + 1})$ where  $V = V_k$ is not a local optimum, the $E^*$ is not in $S_{++}^d$, which means we cannot use the Cholesky decomposition to solve for $V_{k + 1}$ in the next update. 

For the final two equations, we can write $$V_{k + 1}^\top A V_{k + 1} = \left[(VE(V))^{\top} (VE(V)) \right]^{-1} (V E(V))^{\top} B^{-1} H_2 E(V)^{-1},$$ where $H_2 = AVE(V) + \alpha \left(\widetilde{B} V \widetilde{E}(V) \widetilde{G}(V) \widetilde{E}(V) - \widetilde{A}V\widetilde{E}(V) \right)$. However, when $V = V_k$ is not a local optimum, again the right-hand side is not symmetric positive-definite, and so we cannot use the Cholesky decomposition to solve for $V_{k + 1}$  in the next update. 


Note that in order to require $V_{k + 1} \in \stf(p, d)$, the final step of each update rule should project the solution for $V_{k + 1}$ onto $\stf(p, d)$, which can be done via SVD; if $A = U \Sigma V^\top$, then $\pi(A) = U V^\top$. 

Although this cyclic update regime does not immediately lead to a practical fixed-point optimization algorithm, it shows that our loss function has the classical link to a Ricatti equation \eqref{eqn:Ricatti}, indicating more efficient algorithms are possible.

\section{Additional experimental details}
We provide more details about empirical studies in Section \ref{sec:appl}. There are three main purposes of this appendix.
\begin{enumerate}
	\item We provide classification accuracy or prediction MSE from multiple commonly used algorithms since the true function $\varphi$ can be arbitrary in Equation \eqref{eqn:assumption_contrastive}.
	\item We provide the standard deviation that results from random split in cross validation with 10 replicates.
	\item We provide the time  of 8 DR algorithms: PCA, CPCA, LDA, LASSO, SIR CIR, tSNE and UMAP with $d=2$. The run time is based on an personal iMac 2021 with M1 chip.
\end{enumerate}

\subsection{Mouse protein}\label{apdx:mouse}
We first compare the run time for 8 DR algorithms. Note that contrastive models take 552 foreground samples along with 255 background samples so the total sample size is 807, while non-contrastive models take only 552 samples. In contrast, unsupervised methods take 77 input features, i.e., 77 proteins, while supervised method take one additional feature, the response variable $y$ into consideration so $p=77+1=78$. 

\begin{table}[h!]
	\begin{center}
		\caption{Time (seconds) of DR methods on mouse protein data}
		\begin{tabular}{c|c|c|c|c} 
			\hline
			& n & p&d& time \\
			\hline
			
			PCA &     552 &  77   & 2  &  0.01
			\\
			\hline
			CPCA &     807 &  77   & 2& 0.01
			\\
			\hline 
			LDA &       552 &   78  &  2 &   0.02
			\\
			\hline
			LASSO &       552 &  78  &  2  &  0.05
			\\
			\hline
			SIR &       552 &  78 &  2  & 0.05
			\\
			\hline
			
			CIR&   807  & 78  &  2  & 1.16
			\\
			\hline
			tSNE&   552  & 77  &  2  & 0.62
			\\
			\hline
			UMAP&   552  & 77  &  2  & 5.31
			\\
			\hline
		\end{tabular}
		
	\end{center}
\end{table}

We present the classification accuracy by KNN, trees, SVM, boosting and neural network applied to mouse protein data. All models are trained by the MATLAB app: classification learner. The reduced dimension $d=2,3,\cdots,7$ because when $d>7$ the accuracy from almost all methods is close to $1$. We removed the standard deviation for those less than $0.0001$ for simplicity.
\begin{table}[h!]
	\begin{center}
		\caption{Classification accuracy (standard deviation) of KNN for different DR methods and $d$}
		\begin{tabular}{c|c|c|c|c|c|c} 
			\hline
			\diagbox[width=3.5em]{ DR}{d}& 2 & 3&4&5&6&7 \\
			\hline
			raw &0.996(0)  &  0.996(0)&    0.996(0)&    0.995(0)  &  0.994(0)   & 0.995(0)\\
			\hline
			PCA &     0.588(0.01)  &  0.881(0.01)   & 0.945(0)  &  0.971(0)   & 0.974(0) &   0.979(0.01)
			\\
			\hline
			CPCA &     0.614(0.01)  &  0.880(0.01)   & 0.933 (0.01) &  0.959(0.01) &   0.973(0.01)  &  0.978(0.01)
			\\
			\hline 
			LDA &      0.754(0.01) &   0.821(0.01)  &  0.911(0) &   0.966(0)   & 0.966 (0)  & 0.991(0)
			\\
			\hline
			LASSO &      0.464 (0.01) &  0.661(0.01)  &  0.753(0.01)  &  0.859 (0.01)   &0.853(0)   & 0.911(0.01)
			\\
			\hline
			SIR &      0.445(0.01)  &  0.826(0) &   \bf{0.965(0)}   & \bf{0.991(0)}  &  0.974(0)  &  \bf{0.996(0)}
			\\
			\hline
			
			CIR&   \bf{0.772(0)}  & \bf{0.914(0)}  &  0.958(0)  &  0.990(0)   & \bf{0.997(0)} &  0.995(0.01)
			\\
			\hline
		\end{tabular}
		
	\end{center}
\end{table}

\begin{table}[h!]
	\begin{center}
		\caption{Classification accuracy (standard deviation) of trees for different DR methods and $d$}
		\begin{tabular}{c|c|c|c|c|c|c} 
			\hline
			\diagbox[width=3.5em]{ DR}{d}& 2 & 3&4&5&6&7 \\
			\hline
			raw& 0.890(0.01)&0.889(0.01)&0.890(0.01)&0.890(0.01)&0.886(0.01)&0.884(0.01)\\\hline
			PCA&0.427(0.02)&0.703(0.02)&0.805(0.01)&0.819(0.01)&0.846(0.01)&0.850(0.01)\\\hline
			CPCA&0.463(0.02)&0.671(0.02)&0.718(0.01)&0.721(0.02)&0.759(0.01)&0.767(0.01)\\\hline
			LDA&\bf{0.785(0.01)}&\bf{0.879(0.01)}&\bf{0.939(0)}&\bf{0.986(0)}&\bf{0.992(0)}&0.989(0)\\\hline
			LASSO&0.454(0.01)&0.576(0.01)&0.690(0.01)&0.78(0.01)&0.770(0.01)&0.794(0.02)\\\hline
			SIR&0.482(0.01)&0.817(0.01)&0.941(0.01)&0.981(0)&0.983(0)&\bf{0.992(0)}\\\hline
			CIR&0.759(0.01)&0.863(0.01)&0.906(0.01)&0.912(0.01)&0.954(0.01)&0.874(0.01)\\\hline
		\end{tabular}
		
	\end{center}
\end{table}

\begin{table}[h!]
	\begin{center}
		\caption{Classification accuracy (standard deviation) of SVM for different DR methods and $d$}
		\begin{tabular}{c|c|c|c|c|c|c} 
			\hline
			\diagbox[width=3.5em]{ DR}{d}& 2 & 3&4&5&6&7 \\
			\hline
			raw&0.993(0)&0.993(0)&0.993(0)&0.993(0)&0.992(0)&0.994(0)\\\hline
			PCA&0.329(0.01)&0.596(0.01)&0.742(0)&0.803(0.01)&0.825(0)&0.846(0)\\\hline
			CPCA&0.415(0.04)&0.542(0.01)&0.650(0.01)&0.749(0.01)&0.782(0)&0.788(0)\\\hline
			LDA&0.426(0)&0.426 (0)&0.426(0)&0.428(0)&0.428(0)&0.428(0)\\\hline
			LASSO&0.163(0)&0.179(0)&0.350(0.01)&0.505(0.01)&0.502(0.01)&0.506(0.01)\\\hline
			SIR&0.163(0)&0.163(0)&0.163(0)&0.163(0)&0.242(0)&0.246(0)\\\hline
			CIR&\bf{0.749(0)}&\bf{0.849(0)}&\bf{0.882(0)}&\bf{0.876(0)}&\bf{0.942(0)}&\bf{0.924(0)}\\\hline
		\end{tabular}
		
	\end{center}
\end{table}

\begin{table}[h!]
	\begin{center}
		\caption{Classification accuracy (standard deviation) of boosting for different DR methods and $d$}
		\begin{tabular}{c|c|c|c|c|c|c} 
			\hline
			\diagbox[width=3.5em]{ DR}{d}& 2 & 3&4&5&6&7 \\
			\hline
			raw&0.986(0)&0.985(0)&0.984(0)&0.988(0)&0.985(0)&0.986(0)\\\hline
			PCA&0.408(0.01)&0.607(0.01)&0.754(0.01)&0.822(0.01)&0.872(0.01)&0.898(0.01)\\\hline
			CPCA&0.452(0.01)&0.680(0.01)&0.744(0.01)&0.796(0.01)&0.858(0.01)&0.870(0.01)\\\hline
			LDA&\bf{0.790(0.01)}&\bf{0.900(0.01)}&0.950(0)&0.875(0.06)&0.0815(0)&0.08152(0)\\\hline
			LASSO&0.474(0.01)&0.520(0.01)&0.711(0.01)&0.805(0.01)&0.812(0.01)&0.842(0.01)\\\hline
			SIR&0.544(0.01)&0.840(0.01)&\bf{0.964(0.01)}&\bf{0.981(0.03)}&0.953(0.05)&0.082(0)\\\hline
			CIR&0.787(0.01)&0.866(0.01)&0.945(0.04)&0.948(0)&\bf{0.977(0)}&\bf{0.973(0)}\\\hline
		\end{tabular}
		
	\end{center}
\end{table}

\begin{table}[h!]
	\begin{center}
		\caption{Classification accuracy (standard deviation) of neural network classifier for different DR methods and $d$}
		\begin{tabular}{c|c|c|c|c|c|c} 
			\hline
			\diagbox[width=3.5em]{ DR}{d}& 2 & 3&4&5&6&7 \\
			\hline
			raw&0.909(0.04)&0.903(0.06)&0.895(0.03)&0.898(0.03)&0.881(0.06)&0.901(0.04)\\\hline
			PCA&0.466(0.01)&0.798(0.02)&0.90(0.01)&0.924(0.01)&0.934(0.01)&0.943(0.01)\\\hline
			CPCA&0.560(0.01)&0.825(0.01)&0.860(0.01)&0.880(0.01)&0.924(0.01)&0.925(0.01)\\\hline
			LDA&\bf{0.813(0.01)}&\bf{0.877(0.01)}&0.926(0.02)&\bf{0.968(0.02)}&0.976(0.01)&0.965(0.02)\\\hline
			LASSO&0.490(0.01)&0.490(0.03)&0.648(0.02)&0.803(0.02)&0.815(0.02)&0.863(0.01)\\\hline
			SIR&0.55(0.03)&0.835(0.02)&0.953(0.02)&0.964(0.02)&0.964(0.02)&\bf{0.982(0)}\\\hline
			CIR&0.788(0.01)&0.871(0.01)&\bf{0.939(0.01)}&0.951(0)&\bf{0.978(0)}&0.979(0.01)\\\hline
		\end{tabular}
		
	\end{center}
\end{table}

For $d = 2, 3$ and various choices of $\alpha$, we present KNN classification accuracy on this dataset. The small changes in $\alpha$ support the claim that CIR is robust in $\alpha$. For $d = 2$, we provide the corresponding visualizations. \begin{table}[!h]
	\caption{Classification Accuracy (standard deviation) of KNN for CIR, with different $\alpha$ and $d=2,3$.}
	\centering
	\begin{tabular}{c|c|c|c|c|c}
		\hline 
		\diagbox[width=3.5em]{$d$}{$\alpha$} & $5\cdot10^{-5}$ & $10^{-4}$ & $2\cdot10^{-4}$ & $4\cdot10^{-4}$ & $8\cdot10^{-4}$\tabularnewline
		\hline 
		$2$ & 0.740 (0.0237) & 0.775 (0.017) & 0.751 (0.0338) & 0.747 (0.0336) & 0.753 (0.0244)\tabularnewline
		\hline 
		$3$ & 0.874 (0.0241) & 0.860 (0.0122) & 0.869 (0.0270) & 0.867 (0.0306) & 0.855 (0.0265)\tabularnewline
		\hline 
	\end{tabular}
\end{table}

\clearpage

\begin{center}
	\includegraphics[width=\columnwidth]{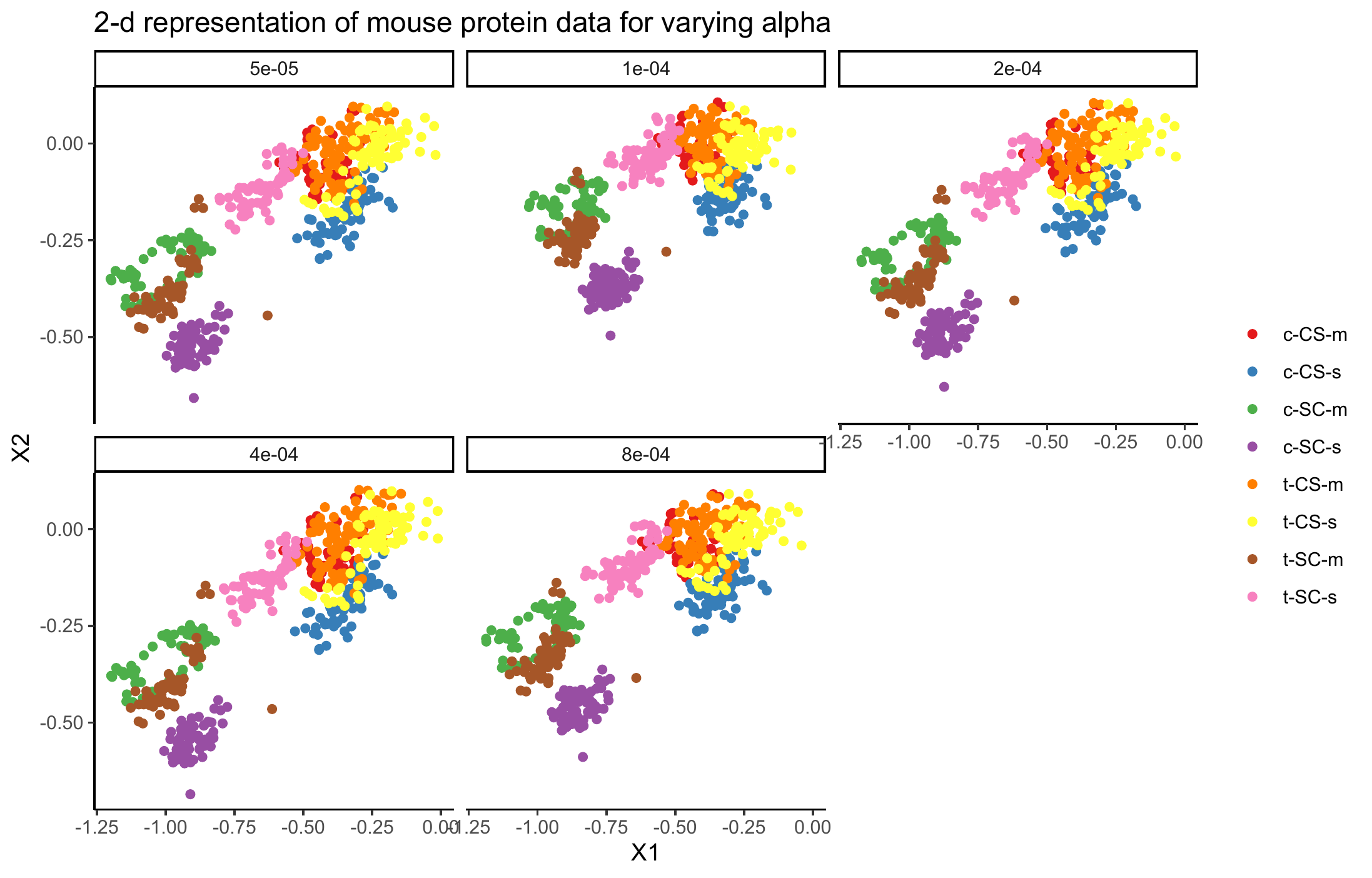} 
	\par\end{center}

\subsection{Single-cell RNA sequencing}\label{apdx:sc}

We first compare the run time for 8 DR algorithms. Note that contrastive models take 3500 foreground samples along with 3500 background samples so the total sample size is 7000, while non-contrastive models take only 3500 samples. In contrast, unsupervised methods take 100 input features, i.e., 100 genes, while supervised method take one additional feature, the response variable $y$ into consideration so $p=100+1=101$. 

\begin{table}[h!]
	\begin{center}
		\caption{Time (seconds) of DR methods on single-cell RNA sequencing data}
		\begin{tabular}{c|c|c|c|c} 
			\hline
			& n & p&d& time \\
			\hline
			
			PCA &     3500&  100   & 2  &  0.03
			\\
			\hline
			CPCA &    7000&  100   & 2& 0.04
			\\
			\hline 
			LDA &       3500 &   101 &  2 &   0.08
			\\
			\hline
			LASSO &       3500 &  101  &  2  &  0.09
			\\
			\hline
			SIR &       3500 &  101 &  2  & 0.06
			\\
			\hline
			
			CIR&   7000  & 101  &  2  & 4.96
			\\
			\hline
			tSNE&   3500  & 100  &  2  & 3.28
			\\
			\hline
			UMAP&   3500  & 100  &  2  & 4.97
			\\
			\hline
		\end{tabular}
		
	\end{center}
\end{table}

We present the classification accuracy by KNN, trees, boosting and neural network applied to single-cell RNA sequencing data. We will not present the accuracy from SVM since the time is much higher than other classifiers, where a single training takes about 20 minutes (we have 10 replicates and 6 options for $d$ so the total time will be 1200 minutes for SVM). All models are trained by the MATLAB app: classification learner. The reduced dimension $d=2,3,\cdots,7$ because when $d>7$ the accuracy remains stable when $d$ keeps increasing. We removed the standard deviation for those less than $0.0001$ for simplicity.


\begin{table}[h!]
	\caption{Classification accuracy (standard deviation) of KNN for different DR methods and $d$}
	\begin{tabular}{c|c|c|c|c|c|c}
		\hline
		\diagbox[width=3.5em]{ DR}{d}& 2 & 3&4&5&6&7 \\
		\hline   raw&0.761(0)&0.761(0)&0.762(0)&0.762(0)&0.762(0)&0.762(0)\\\hline
		PCA&0.411(0)&0.577(0)&0.588(0)&0.635(0)&0.676(0)&0.704(0)\\
		\hline
		CPCA&0.412(0)&0.575(0)&0.587(0)&0.635(0)&0.676(0)&0.703(0)\\\hline
		LDA&0.659(0)&0.776(0)&\bf{0.844(0)}&0.844(0)&0.862(0)&\bf{0.877(0)}\\\hline
		LASSO&0.534(0)&0.577(0)&0.642(0)&0.697(0)&0.696(0)&0.697(0)\\\hline
		SIR&0.481(0.01)&0.570(0.01)&0.661(0)&0.690(0)&0.832(0)&0.871(0)\\\hline
		CIR&\bf{0.711(0)}&\bf{0.823(0)}&0.833(0)&\bf{0.866(0)}&\bf{0.864(0)}&0.876(0)\\\hline
	\end{tabular}
	
\end{table}

\begin{table}[h!]
	\begin{center}
		\caption{Classification accuracy (standard deviation) of Trees for different DR methods and $d$}
		\begin{tabular}{c|c|c|c|c|c|c} 
			\hline
			\diagbox[width=3.5em]{ DR}{d}& 2 & 3&4&5&6&7\\
			\hline
			raw&0.753(0.01)&0.758(0.01)&0.756(0.01)&0.756(0)&0.754(0.01)&0.755(0)\\\hline
			PCA&0.450(0.01)&0.590(0.01)&0.633(0.01)&0.653(0.01)&0.710(0)&0.731(0.01)\\\hline
			CPCA&0.448(0.01)&0.591(0)&0.634(0)&0.656(0.01)&0.716(0)&0.729(0)\\\hline
			LDA&0.676(0)&0.776(0)&\bf{0.839(0)}&0.835(0)&0.840(0)&0.862(0)\\\hline
			LASSO&0.555(0.01)&0.585(0.01)&0.634(0.01)&0.692(0.01)&0.689(0.01)&0.690(0)\\\hline
			SIR&0.510(0.01)&0.603(0)&0.668(0.01)&0.685(0.01)&0.825(0.01)&0.859(0)\\\hline
			CIR&\bf{0.721(0)}&\bf{0.821(0.01)}&0.827(0)&\bf{0.845(0)}&\bf{0.853(0.01)}&\bf{0.868(0)}\\\hline
		\end{tabular}
		
	\end{center}
\end{table}

\begin{table}[h!]
	\begin{center}
		\caption{Classification accuracy (standard deviation) of boosting for different DR methods and $d$}
		\begin{tabular}{c|c|c|c|c|c|c} 
			\hline
			\diagbox[width=3.5em]{ DR}{d}& 2 & 3&4&5&6&7\\
			\hline
			raw&0.826(0)&0.827(0)&0.826(0)&0.828(0)&0.826(0)&0.827(0)\\\hline
			PCA&0.540(0)&0.689(0)&0.714(0)&0.722(0)&0.777(0)&0.794(0)\\\hline
			CPCA&0.539(0)&0.690(0)&0.714(0)&0.723(0)&0.779(0)&0.795(0)\\\hline
			LDA&0.747(0)&0.832(0)&\bf{0.872(0)}&0.877(0)&0.876(0)&0.895(0)\\\hline
			LASSO&0.662(0)&0.692(0)&0.711(0)&0.769(0)&0.768(0)&0.768(0)\\\hline
			SIR&0.624(0)&0.688(0)&0.7419(0)&0.754(0)&0.864(0)&0.889(0)\\\hline
			CIR&\bf{0.786(0)}&\bf{0.858(0)}&0.868(0)&\bf{0.882(0)}&\bf{0.889(0)}&\bf{0.896(0)}\\\hline
		\end{tabular}
		
	\end{center}
\end{table}


\begin{table}[h!]
	\begin{center}
		\caption{Classification accuracy (standard deviation) of neural network classifier for different DR methods and $d$}
		\begin{tabular}{c|c|c|c|c|c|c} 
			\hline
			\diagbox[width=3.5em]{ DR}{d}& 2 & 3&4&5&6&7 \\
			\hline
			raw&0.870(0.01)&0.859(0.03)&0.870(0.0220)&0.867(0.02)&0.870(0.01)&0.870(0.02)\\\hline
			PCA&0.514(0.01)&0.671(0.02)&0.683(0.04)&0.725(0.02)&0.784(0.02)&0.794(0.03)\\\hline
			CPCA&0.506(0.01)&0.678(0.01)&0.705(0.03)&0.758(0.01)&0.782(0.03)&0.797(0.03)\\\hline
			LDA&0.748(0)&0.841(0)&\bf{0.886(0)}&0.886(0)&0.886(0)&0.904(0)\\\hline
			LASSO&0.661(0)&0.697(0)&0.757(0)&0.790(0)&0.789(0)&0.789(0)\\\hline
			SIR&0.628(0)&0.701(0)&0.754(0)&0.771(0)&0.878(0)&0.898(0)\\\hline
			CIR&\bf{0.796(0)}&\bf{0.880(0)}&0.881(0)&\bf{0.890(0)}&\bf{0.903(0)}&\bf{0.905(0)}\\\hline
		\end{tabular}
		
	\end{center}
\end{table}

For $d = 2, 3$, $p = 100, 200, 300, 400, 500$, and various choices of $\alpha$, we present KNN classification accuracy on this dataset. The small changes in both $\alpha$ and $p$ support the claim that CIR is robust in $\alpha$ and $p$. For $d = 2$, we provide the corresponding visualizations.

\begin{table}[!h]
	\caption{Classification Accuracy (standard deviation) of KNN for CIR, with different $\alpha$, $p$ and $d=2$.}
	\centering
	\begin{tabular}{c|c|c|c|c|c}
		\hline 
		\diagbox[width=3.5em]{$p$}{$\alpha$} & $0.375$ & $0.75$ & $1.5$ & $3$ & $6$\tabularnewline
		\hline 
		100 & 0.747 (0.015) & 0.741 (0.019) & 0.742 (0.026) & 0.722 (0.011) & 0.707 (0.018)\tabularnewline
		\hline 
		200 & 0.766 (0.006) & 0.771 (0.012) & 0.764 (0.019) & 0.741 (0.007) & 0.718 (0.008)\tabularnewline
		\hline 
		300 & 0.777 (0.006) & 0.770 (0.013) & 0.773 (0.006) & 0.756 (0.007) & 0.720 (0.006)\tabularnewline
		\hline
		400 & 0.779 (0.011) & 0.775 (0.006) & 0.777 (0.015) & 0.763 (0.015) & 0.724 (0.008)\tabularnewline
		\hline
		500 & 0.785 (0.008) & 0.783 (0.015) & 0.784 (0.016) & 0.762 (0.017) & 0.727 (0.012)\tabularnewline
		\hline
	\end{tabular}
\end{table}

\begin{table}[!h]
	\caption{Classification Accuracy (standard deviation) of KNN for CIR, with different $\alpha$, $p$ and $d=3$.}
	\centering
	\begin{tabular}{c|c|c|c|c|c}
		\hline 
		\diagbox[width=3.5em]{$p$}{$\alpha$} & $0.375$ & $0.75$ & $1.5$ & $3$ & $6$\tabularnewline
		\hline 
		100 & 0.810 (0.010) & 0.813 (0.005) & 0.810 (0.010) & 0.794 (0.012) & 0.760 (0.008)\tabularnewline
		\hline 
		200 & 0.841 (0.012) & 0.854 (0.003) & 0.842 (0.011) & 0.828 (0.006) & 0.808 (0.008)\tabularnewline
		\hline 
		300 & 0.865 (0.009) & 0.854 (0.005) & 0.862 (0.013) & 0.838 (0.004) & 0.806 (0.003)\tabularnewline
		\hline
		400 & 0.861 (0.007) & 0.857 (0.008) & 0.850 (0.013) & 0.834 (0.005) & 0.812 (0.005)\tabularnewline
		\hline
		500 & 0.867 (0.007) & 0.858 (0.005) & 0.853 (0.011) & 0.848 (0.013) & 0.816 (0.007)\tabularnewline
		\hline
	\end{tabular}
\end{table}

\begin{figure}[!h]
	\centering
	\includegraphics[width=\columnwidth]
	{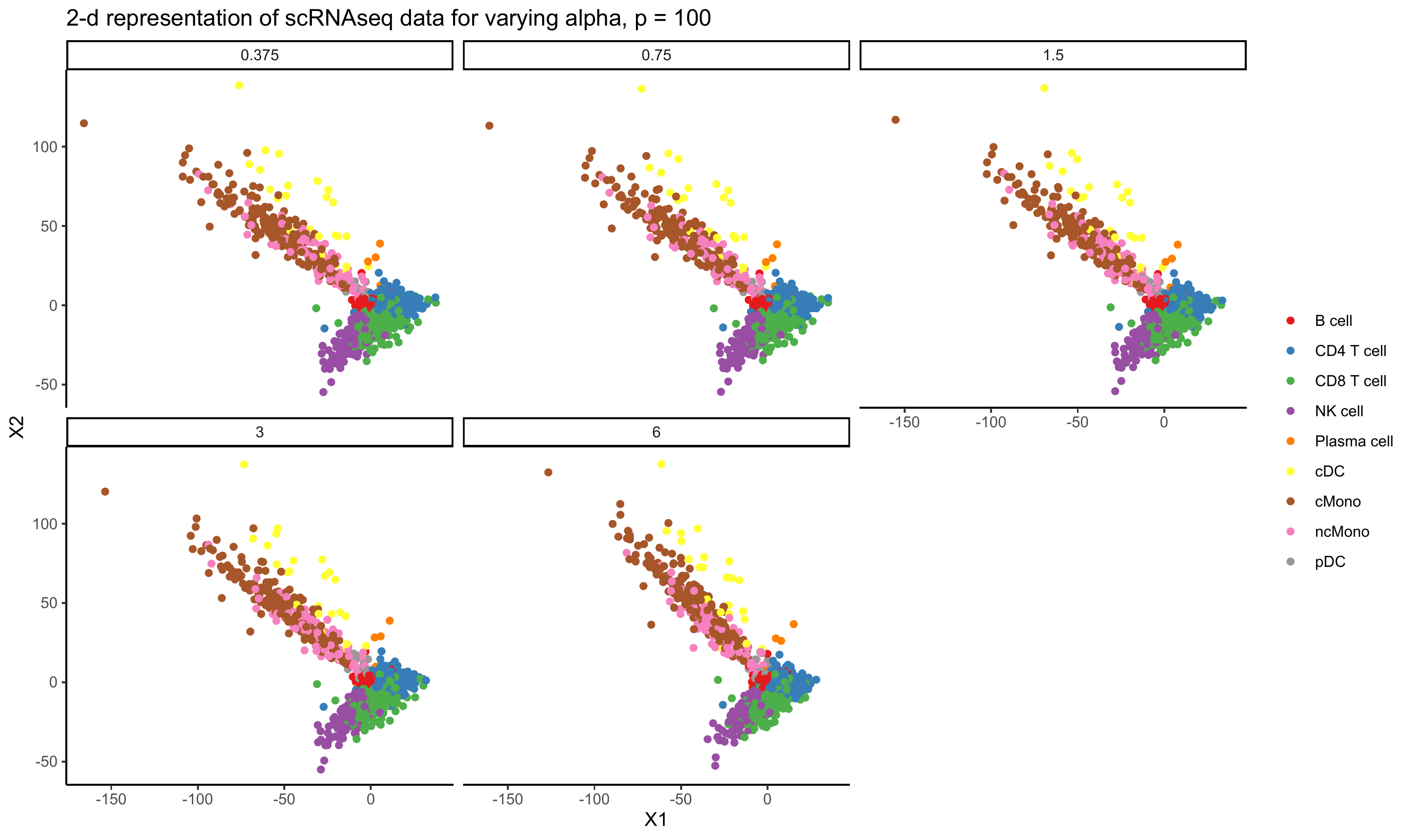}
\end{figure}
\begin{figure}[!h]
	\centering
	\includegraphics[width=\columnwidth]{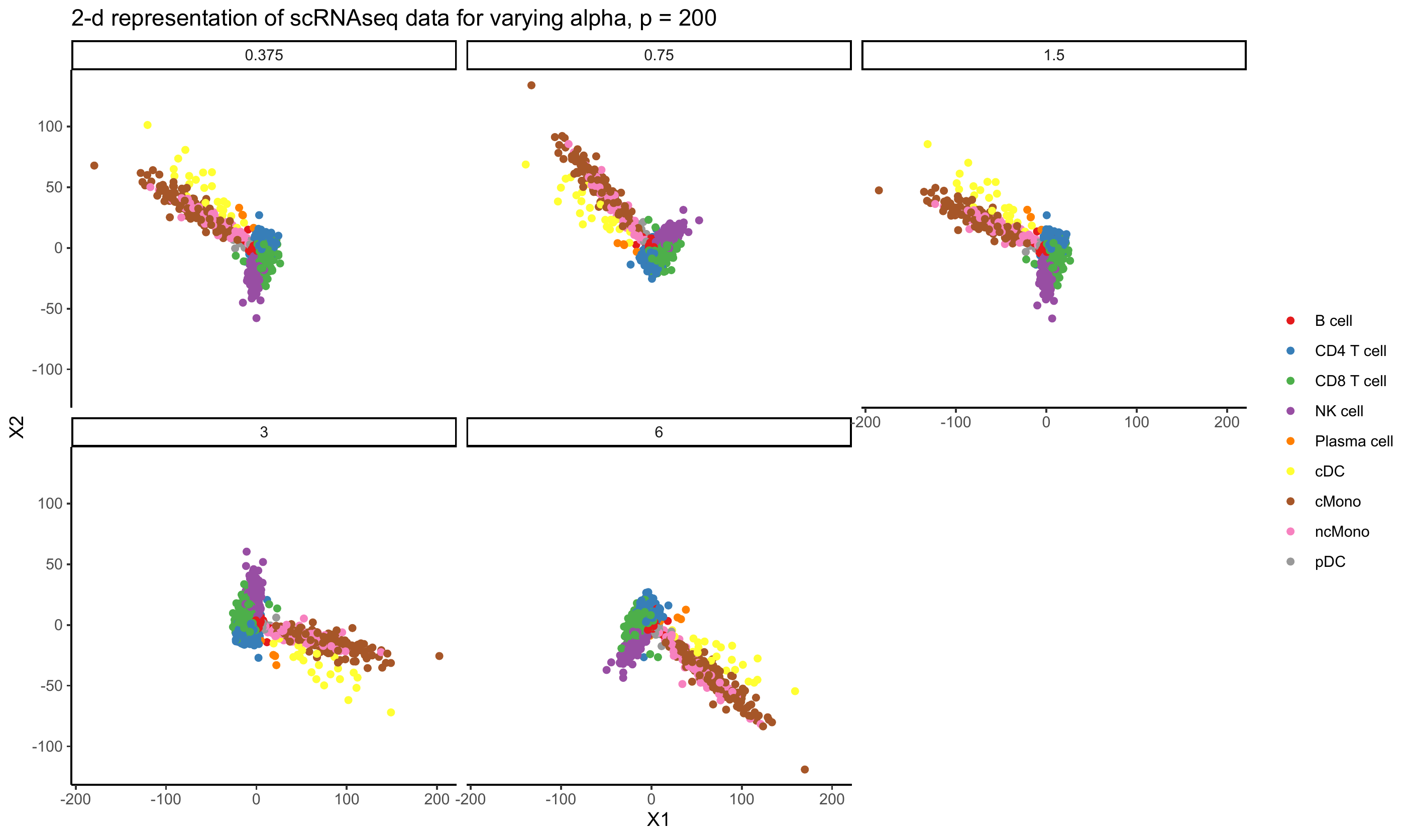}
\end{figure}
\begin{figure}[!h]
	\centering
	\includegraphics[width=\columnwidth]{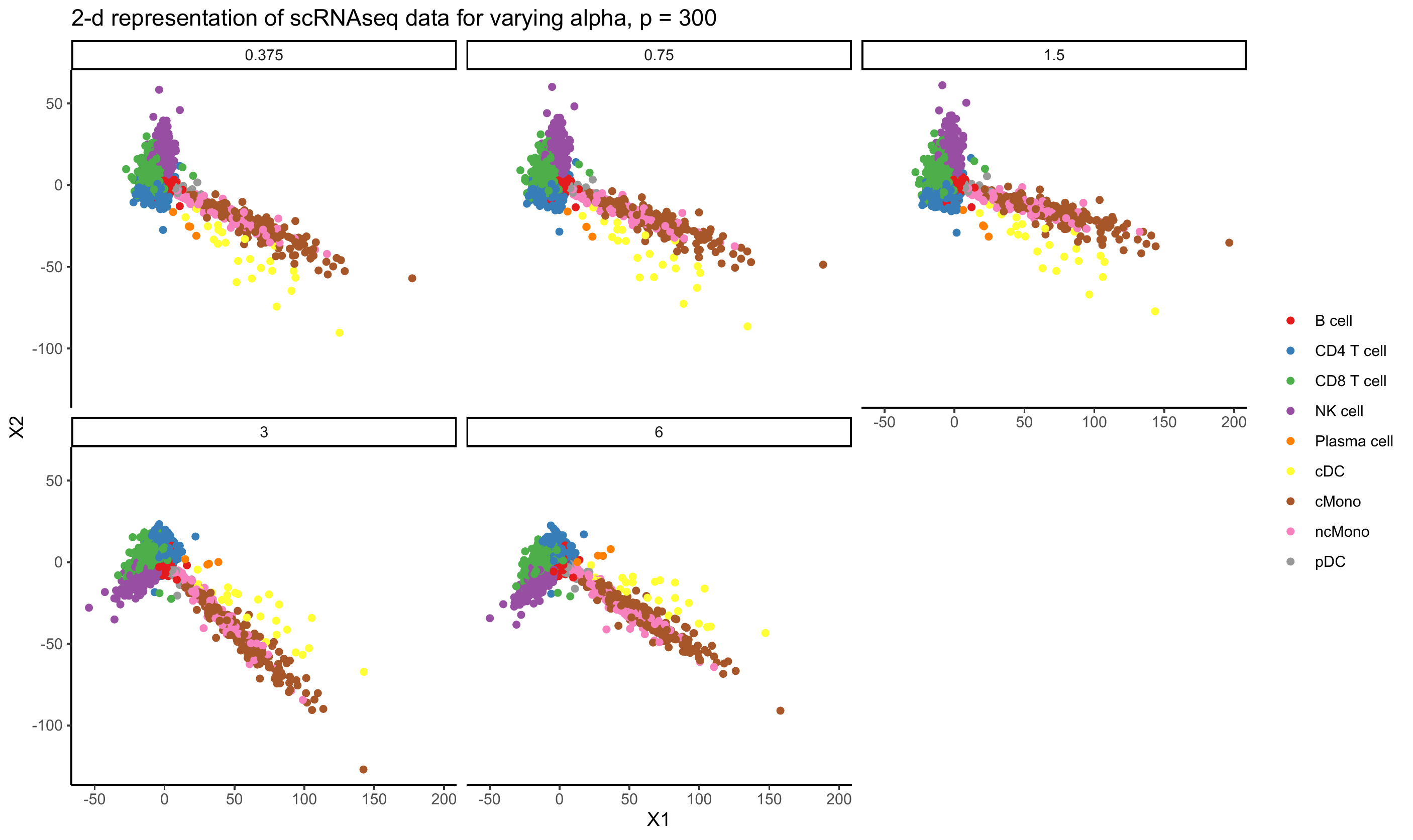}
\end{figure}
\begin{figure}[!h]
	\centering
	\includegraphics[width=\columnwidth]{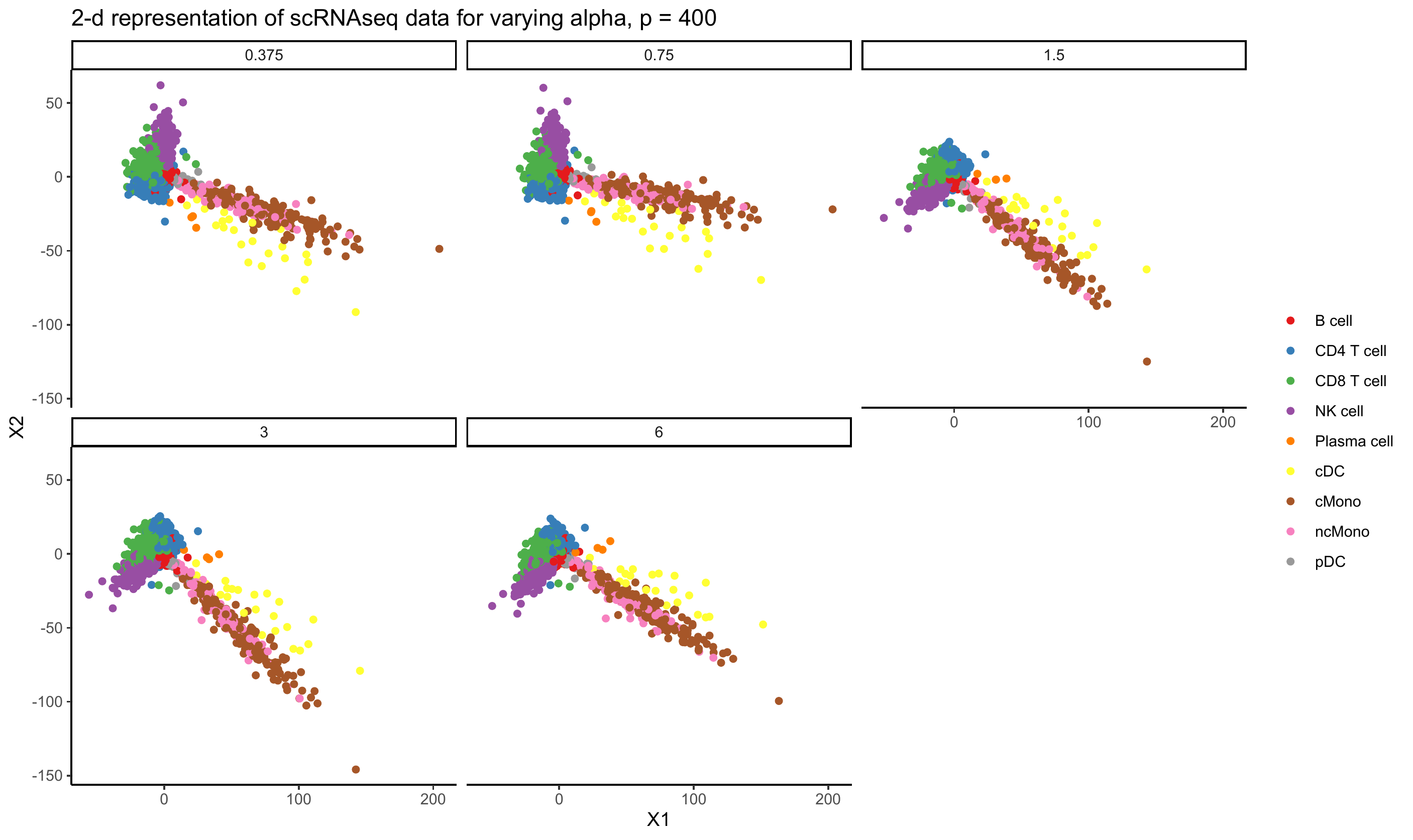}
\end{figure}
\begin{figure}[!h]
	\centering
	\includegraphics[width=\columnwidth]{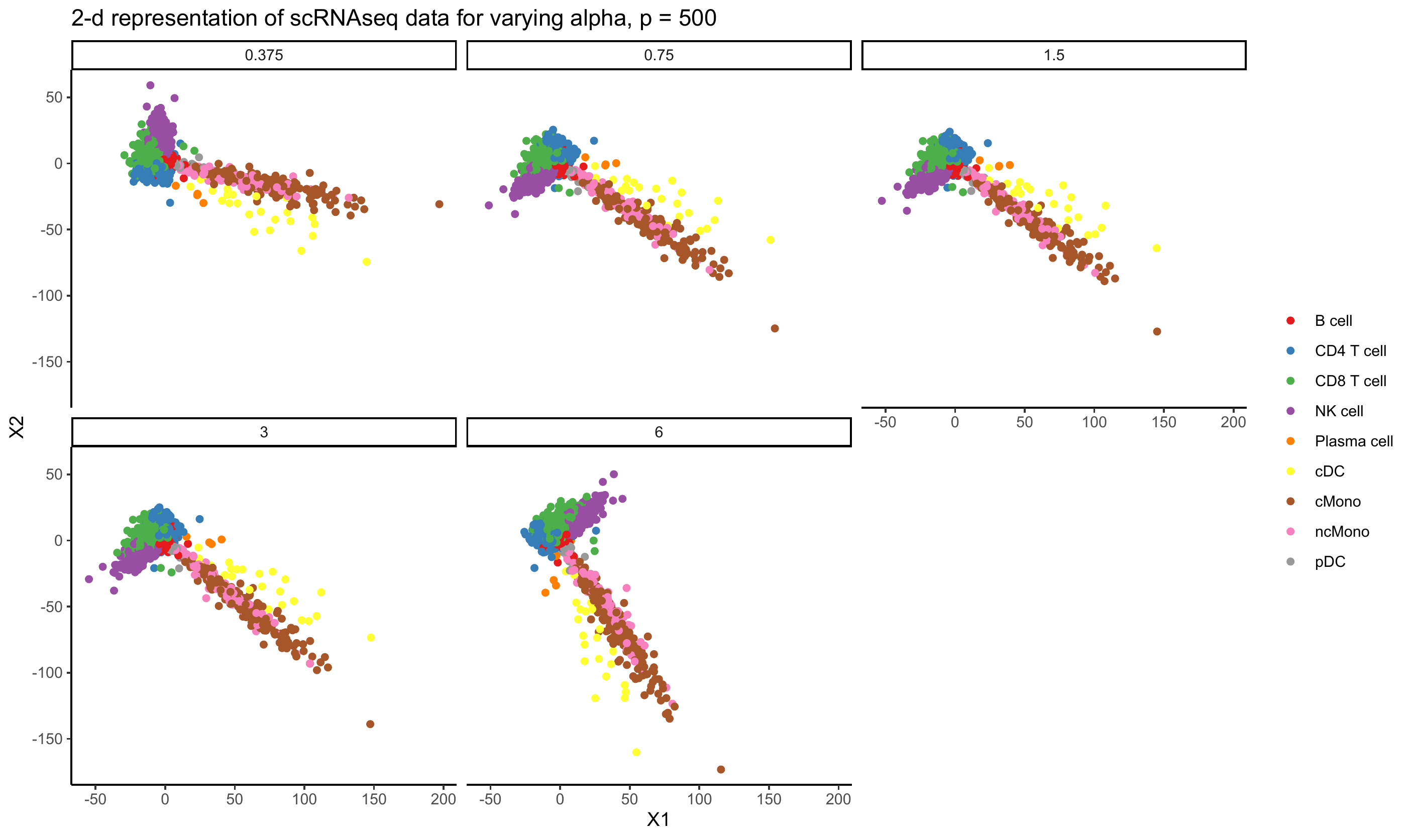}
\end{figure}

\subsection{Plasma retinol}\label{apdx:plasma}
We first compare the run time for 8 DR algorithms. Note that contrastive models take 315 foreground samples along with 315 background samples so the total sample size is 630, while non-contrastive models take only 315 samples. In contrast, unsupervised methods take 12 input features, while supervised method take one additional feature, the response variable $y$ into consideration so $p=12+1=13$. 

\begin{table}[h!]
	\begin{center}
		\caption{Time (seconds) of DR methods on plasma retinol data}
		\begin{tabular}{c|c|c|c|c} 
			\hline
			& n & p&d& time \\
			\hline
			
			PCA &     315 &  12   & 2  &  0.02
			\\
			\hline
			CPCA &   630 &  12   & 2& 0.02
			\\
			\hline 
			LDA &       315 &  13  &  2 &   0.03
			\\
			\hline
			LASSO &       315&  13  &  2  &  0.05
			\\
			\hline
			SIR &       552 &  13&  2  & 0.05
			\\
			\hline
			
			CIR&   630  & 13 &  2  & 0.21
			\\
			\hline
			tSNE&   315  & 12  &  2  & 0.39
			\\
			\hline
			UMAP&   315  & 12  &  2  & 4.16
			\\
			\hline
		\end{tabular}
		
	\end{center}
\end{table}

We present the regression MSE by linear regression, trees, SVM, GP and neural network applied to plasma retinol data. All models are trained by the MATLAB app: regression learner. The reduced dimension $d=1,2,\cdots,8$ because when $d>8$ MSE will not change significantly. We note that we removed the standard deviation for those less than $10$ for simplicity.

Note that for table \ref{tab:linreg}, the standard deviation of every entry is less than $10^{-4}$.

\begin{table}[h!]
	\centering
	\caption{\label{tab:linreg} Prediction MSE of linear regression for different DR methods and $d$}
	\begin{tabular}{c|c|c|c|c|c|c|c|c} 
		\hline
		\diagbox[width=3.5em]{ DR}{d}& 1&2 & 3&4&5&6&7&8\\
		\hline
		raw&28112&28112&28112&28112&28112&28112&28112&28112\\\hline
		PCA&31997&31754&31850&31324&31282&31304&31033&29572\\\hline
		CPCA&31997&31754&31850&31324&31282&31304&31033&29572\\\hline
		LDA&32713&32548&32523&31899&31279&29354&29317&29392\\\hline
		LASSO&31725&\bf{30170}&29837&29023&27987&27979&27882&27949\\\hline
		SIR&33595&33108&\bf{27349}&\bf{27720}&\bf{27804}&\bf{27658}&\bf{27733}&\bf{27764}\\\hline
		CIR&\bf{30139}&30216&27833&\bf{27720}&\bf{27804}&\bf{27658}&\bf{27733}&\bf{27764}\\\hline
	\end{tabular}
	
\end{table}


\begin{table}[h!]
	\centering   \caption{Prediction MSE$/100$ (standard deviation$/100$) of regression tree for different DR methods and $d$}
	\begin{tabular}{c|c|c|c|c|c|c|c|c} 
		\hline
		\diagbox[width=3.5em]{ DR}{d}& 1&2 & 3&4&5&6&7&8\\
		\hline
		raw&515(32)&481(30)&476(37)&461(34)&462(42)&443(46)&469(34)&459(38)\\\hline
		PCA&519(42)&537(36)&532(78)&580(67)&592(55)&594(67)&581(50)&576(47)\\\hline
		CPCA&513(40)&555(64)&561(55)&565(59)&564(40)&564(39)&561(27)&529(58)\\\hline
		LDA&476(22)&538(21)&540(40)&584(58)&490(50)&515(75)&508(48)&512(69)\\\hline
		LASSO&498(25)&\bf{472(40)}&\bf{463(50)}&472(34)&506(42)&502(63)&488(42)&488(42)\\\hline
		SIR&\bf{457(29)}&529(32)&486(54)&\bf{441(31)}&479(43)&453(35)&\bf{435(44)}&\bf{456(42)}\\\hline
		CIR&515(32)&481(30)&476(37)&461(34)&\bf{462(42)}&\bf{443(46)}&469(34)&459(38)\\\hline
	\end{tabular}
	
\end{table}

\begin{table}[h!]
	\centering   \caption{Prediction MSE$/100$ (standard deviation$/100$) of SVM for different DR methods and $d$}
	\begin{tabular}{c|c|c|c|c|c|c|c|c} 
		\hline
		\diagbox[width=3.5em]{ DR}{d}& 1&2 & 3&4&5&6&7&8\\
		\hline
		raw&323(2)&322(2)&309(2)&307(1)&309(2)&305(2)&307(2)&307(1)\\\hline
		PCA&352(4)&350(6)&349(9)&349(10)&350(12)&344(5)&345(10)&332(8)\\\hline
		CPCA&353(4)&344(4)&351(11)&348(8)&347(9)&348(16)&346(7)&334(11)\\\hline
		LDA&352(2)&349(2)&350(2)&339(2)&333(2)&327(2)&326(2)&326(1)\\\hline
		LASSO&342(1)&333(2)&337(5)&321(5)&316(5)&311(3)&311(6)&312(5)\\\hline
		SIR&361(1)&362(1)&\bf{304(1)}&308(1)&\bf{309(2)}&306(2)&\bf{306(2)}&\bf{306(2)}\\\hline
		CIR&\bf{323(2)}&\bf{322(2)}&309(2)&\bf{307(1)}&309(2)&\bf{305(2)}&307(2)&307(1)\\\hline
	\end{tabular}
	
\end{table}

\begin{table}[!ht]
	\centering
	\caption{Prediction MSE$/100$ (standard deviation$/100$) of GP regression for different DR methods and $d$}
	\begin{tabular}{c|c|c|c|c|c|c|c|c} 
		\hline
		\diagbox[width=3.5em]{ DR}{d}& 1&2 & 3&4&5&6&7&8\\
		\hline
		raw&311(3)&341(12)&299(8)&298(10)&296(10)&292(6)&295(7)&291(7)\\\hline
		PCA&328(4)&323(2)&324(4)&324(2)&323(5)&323(3)&326(6)&325(4)\\\hline
		CPCA&330(5)&322(2)&324(3)&323(3)&322(2)&325(6)&324(3)&324(2)\\\hline
		LDA&337(3)&345(3)&341(6)&334(3)&338(5)&330(8)&327(7)&330(8)\\\hline
		LASSO&334(6)&\bf{310(9)}&335(4)&336(11)&335(10)&329(4)&328(5)&330(3)\\\hline
		SIR&336(1)&342(4)&\bf{285(6)}&299(11)&\bf{295(6)}&296(7)&\bf{293(7)}&\bf{289(5)}\\\hline
		CIR&\bf{311(3)}&341(12)&299(8)&\bf{298(10)}&296(10)&\bf{292(6)}&295(7)&291(7)\\\hline
	\end{tabular}
	
\end{table}



For table \ref{tab:nnregresion} (neural network regression), according to the baseline MSE from raw data, we replace the MSEs greater than $100000$ or standard deviation greater than $10000$ by $*$, which might result from over-fitting. 
\begin{table}[!ht]
	\centering
	\caption{\label{tab:nnregresion} Prediction MSE$/100$ (standard deviation$/100$) of neural network regression for different DR methods and $d$}
	\begin{tabular}{c|c|c|c|c|c|c|c|c} 
		\hline
		\diagbox[width=3.8em]{ DR}{d}& 1&2 & 3&4&5&6&7&8\\
		\hline
		raw&319(5)&347(14)&516(*)&390(53)&410(53)&560(*)&595(*)&645(*)\\\hline

		PCA & *&*&*&*&*&*&*&*\\ \hline
		CPCA & *&*&*&*&*&*&*&*\\ \hline
		
		LDA&345(9)&349(13)&\bf{378(16)}&392(21)&398(38)&\bf{380(21)}&\bf{399(29)}&\bf{479(75)}\\\hline

		
		LASSO&344(32)&\bf{324(9)}&*&*&*&*&*&*\\\hline
		
		
		SIR&343(5)&410(66)&605(*)&\bf{379(36)}&\bf{405(55)}&465(59)&513(*)&667(*)\\\hline
		
		
		CIR&\bf{319(5)}&347(14)&516(*)&390(53)&410(53)&560(*)&595(*)&645(*)\\\hline
		
	\end{tabular}
	
\end{table}

\end{document}